\documentclass [final] {IEEEtran}
\IEEEoverridecommandlockouts
\usepackage[colorlinks,urlcolor=blue,linkcolor=blue,citecolor=blue]{hyperref}
\usepackage{color,array}
\usepackage{graphicx}
\usepackage{algorithm}
\usepackage{algorithmic}
\makeatother
\usepackage[utf8]{inputenc}
\usepackage[table]{xcolor}
\usepackage{tabularx,booktabs}
\newcolumntype{C}{>{\centering\arraybackslash}X} 
\usepackage{cite}
\usepackage{amsmath}
\usepackage{amsmath,amssymb,amsfonts}
\usepackage{algorithmic}
\usepackage{graphicx}
\usepackage{textcomp}
\usepackage{xcolor}
\usepackage{hyperref}
\usepackage{caption}
\usepackage{pifont}
\usepackage{xcolor}
\usepackage{rotating}
\usepackage{rotfloat}
\usepackage{algorithm}
\usepackage{algorithmic}
\usepackage{gensymb}
\usepackage{textcomp}
\usepackage{placeins}
\usepackage{balance}
\usepackage{subfigure}
\usepackage{booktabs}
\usepackage{tabularx}
\usepackage{array}
\usepackage{pbox}
\usepackage{longtable}
\usepackage{booktabs}
\def\BibTeX{{\rm B\kern-.05em{\sc i\kern-.025em b}\kern-.08em
    T\kern-.1667em\lower.7ex\hbox{E}\kern-.125emX}}

\makeatletter
\def\endthebibliography{%
  \def\@noitemerr{\@latex@warning{Empty `thebibliography' environment}}%
  \endlist
}
\makeatother

\begin{document}

\title{Automated Speaker Independent Visual Speech Recognition: A Comprehensive Survey} 
\author{Praneeth Nemani, Ghanta Sai Krishna, Kundrapu Supriya, Santosh Kumar \IEEEmembership{Member, IEEE}

}

%\author{Praneeth Nemani, Ghanta Sai Krishna, Nikhil Ramisetty, B Digvijay Sri Sai, Santosh Kumar
%}
%Video-based
%Person Re-identification
%\author{}
%\author{\IEEEauthorblockN{Praneeth Nemani, Nikhil Ramisetty, B. Digvijay Sri Sai}
%\IEEEauthorblockA{{\{praneeth19100, ramisetty19100, bodicherla19100, santosh\}@iiitnr.edu.in}\\
%Department of Computer Science and Engineering\\
%International Institute of Information Technology, Naya Raipur, Chhattisgarh, %India}

%\and
%\and
%\and
%\and
%}

\maketitle
\begin{abstract}
Speaker-independent visual speech recognition (VSR) is a complex task that involves identifying spoken words or phrases from video recordings of a speaker's facial movements. Decoding the intricate visual dynamics of a speaker's mouth in a high-dimensional space is a significant challenge in this field. To address this challenge, researchers have employed advanced techniques that enable machines to recognize human speech through visual cues automatically. Over the years, there has been a considerable amount of research in the field of VSR involving different algorithms and datasets to evaluate system performance. These efforts have resulted in significant progress in developing effective VSR models, creating new opportunities for further research in this area. This survey provides a detailed examination of the progression of VSR over the past three decades, with a particular emphasis on the transition from speaker-dependent to speaker-independent systems. We also provide a comprehensive overview of the various datasets used in VSR research and the preprocessing techniques employed to achieve speaker independence. The survey covers the works published from 1990 to 2023, thoroughly analyzing each work and comparing them on various parameters. This survey provides an in-depth analysis of speaker-independent VSR systems evolution from 1990 to 2023. It outlines the development of VSR systems over time and highlights the need to develop end-to-end pipelines for speaker-independent VSR. The pictorial representation offers a clear and concise overview of the techniques used in speaker-independent VSR, thereby aiding in the comprehension and analysis of the various methodologies. The survey also highlights the strengths and limitations of each technique and provides insights into developing novel approaches for analyzing visual speech cues. Overall, This comprehensive review provides insights into the current state-of-the-art speaker-independent VSR and highlights potential areas for future research.
\end{abstract}


\section{Introduction}
\textbf{Speech recognition} is a process of converting spoken language into written text, which can be used in various applications such as voice commands, transcription, and translation \cite{gaikwad2010review, reddy1976speech}. It involves the analysis of the acoustic features of speech, which can be either audio signals or visual cues like lip movements. \textbf{Audio Analysis} and \textbf{Visual Speech Recognition (VSR)} are two main approaches to speech recognition that share some commonalities. Audio analysis involves the analysis of audio signals to extract speech information. In speech recognition systems, audio analysis is performed by converting the analog audio signal to a digital representation and then applying various signal processing techniques to extract the relevant features of the speech signal. These features are then used to recognize the speech content and convert it to text or some other form of output. \textbf{Automated VSR} is a form of speech recognition technology that relies on the analysis of lip movements and utterances captured on videotape to recognize spoken words or sentences. Unlike audio-based speech recognition, VSR works by analyzing visual information to classify what is being said or spoken. The process of reading lip movements and articulations alone is quite challenging for a novice, as it is inherently ambiguous at different word levels due to homophones - different characters that produce exactly similar lip sequences, such as 'p' and 'b'. However, these ambiguities have been addressed to some extent by using the context of neighboring words in a sentence or a language model to help disambiguate the spoken words. 

VSR has a rich history, dating back to the 16th century when Ponce de León used lip movements to teach the deaf. Today, VSR has become an important tool in several applications, including smartphone-based systems and models that convert silent lip movements into the text to help individuals with hearing impairments communicate in noisy environments. However, despite the significant advancements made in VSR technology over the years, there are still instances where audio analysis outperforms VSR. Audio analysis has proven to be more reliable in situations where visual cues for speech recognition are limited or ambiguous, such as in low-light conditions or when the speaker's face is partially or fully obscured. Similarly, audio analysis is more effective when the speaker's lip movements are not easily distinguishable or when the speaker is not speaking in the viewer's native language. In such cases, relying solely on VSR technology would be less effective. Furthermore, VSR may be more susceptible to errors caused by variations in the speaker's speech patterns and visual cues, such as changes in facial expression or head movement. Such variations can make it challenging for the VSR system to recognize spoken words accurately. On the other hand, audio analysis may be less affected by such variations, making it more reliable for speech recognition in certain contexts.

\subsection{Exploring the roots of VSR: Traditional Methods}
Traditional methods of automatic VSR refer to the use of different techniques and architectures for extracting features and classifying visual speech data. The feature extraction process is mainly based on pixel or shape-based methods or a combination of both. These extracted features are then transformed into the desired format by using mathematical functions like DCT \cite{ahmed1974discrete}, PCA \cite{abdi2010principal}, and DWT \cite{edwards1991discrete}. Once the features have been extracted and transformed, classification is performed using architectures such as HMM and SVMs. However, traditional methods of VSR have certain limitations that can affect their accuracy and effectiveness. One major limitation is their reliance on the quality of the input video data, which can be affected by factors such as lighting, facial expressions, and head movements. These factors can cause variations in the visual speech data, making it difficult for traditional VSR methods to recognize speech accurately. Additionally, traditional methods may not be able to handle variations in pronunciation and dialects, as they are trained on a limited set of data, making them less effective for recognizing speech from diverse speakers. Another limitation of traditional VSR methods is their computational complexity, which can make them slow and resource-intensive. The feature extraction process and classification algorithms require significant processing power, which can limit their use in real-time applications, such as in noisy environments where rapid speech recognition is required. In conclusion, traditional methods of VSR have limitations that can affect their accuracy and effectiveness. The reliance on the quality of input video data, the inability to handle variations in pronunciation and dialects, and computational complexity are some of the significant limitations of traditional VSR methods. Therefore, newer approaches, such as deep learning-based VSR, have been developed to address these limitations and provide more accurate and efficient speech recognition.

\subsection{Revolutionizing VSR with Deep Learning}
In recent years, VSR systems have benefited from the use of Deep Learning (DL) networks for feature extraction and classification. Various types of networks, such as Convolutional Neural Networks (CNNs), Feed-forward Networks, and Autoencoders, have been employed in the front end of lip-reading systems to extract significant features and learn spatial and temporal information. In addition, sequence-processing networks like Recurrent Neural Networks (RNNs) have been commonly used in the backend to classify sequential speech such as words or sentences. Recently, attention-based transformers and temporal convolutional networks have also been adopted as alternatives to RNNs for classification. However, training deep learning models for VSR can be computationally complex and require large amounts of data. To address this issue, transfer learning has been applied to improve the performance of VSR systems. Transfer learning involves using a pre-trained model to improve predictions on a new but similar problem. This approach can reduce training time and data requirements while improving neural network performance. Transfer learning is particularly useful when there is insufficient data to train a model from scratch, and a pre-trained model is available on a related problem with a large volume of data.

\begin{figure}[ht]
    \centering
    \includegraphics[width = \linewidth]{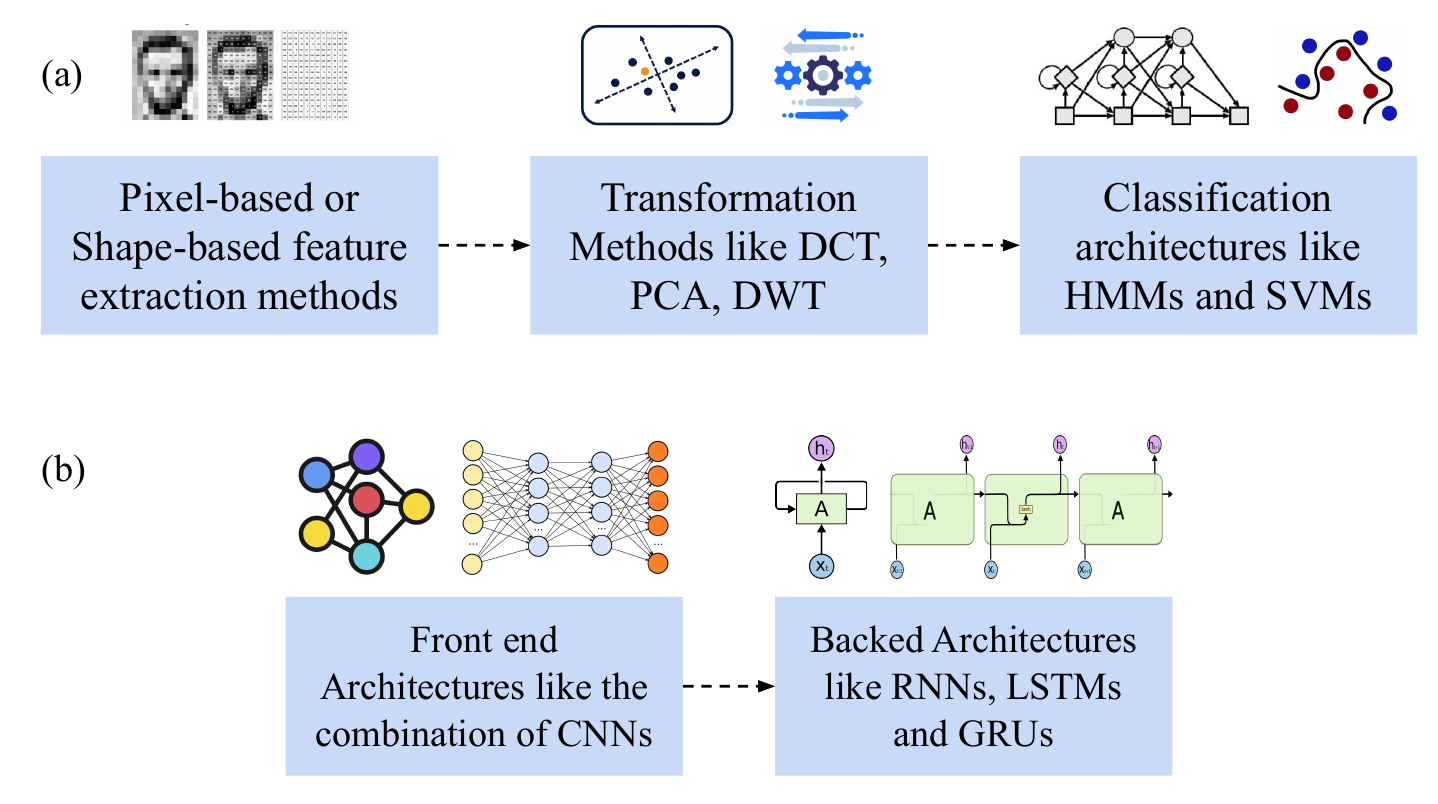}
    \caption{(a) Outline of Traditional VSR Architectures, (b) DL Architectures}
    \label{fig:TADLA}
\end{figure}

Traditional VSR approaches rely on hand-crafted feature extraction techniques that require prior knowledge and expertise, resulting in limited accuracy and susceptibility to lighting, background, and pose variations. For instance, if the selected features are not appropriate for a particular speech signal, the performance of the system may suffer. Moreover, traditional methods often fail to capture the underlying patterns in the data due to the limited representation power of the hand-crafted features. In contrast, DL-based VSR methodologies are more robust to these variations and have shown significant improvements in accuracy. Although traditional VSR methods are computationally less intensive and require less training data than DL-based VSR methods, their ability to handle complex and noisy data is limited, and their performance heavily depends on the quality of the hand-crafted features. On the other hand, DL-based VSR methods require a large amount of data and computational resources for training, but they can handle complex and noisy data, learn more abstract features, and achieve higher accuracy. Thus, the choice of VSR methodology depends on the specific application and the available resources. In conclusion, traditional and DL-based VSR methods have advantages and limitations. The choice of the VSR method depends on the specific application requirements and available resources. Fig. \ref{fig:TADLA} depicts the outline of both traditional and DL methodologies for VSR.  

\subsection{Speaker Independent VSR}
\textbf{Speaker-independent VSR} is a type of VSR system that is trained to recognize speech patterns and lip movements from various speakers without requiring prior training on the specific speaker. This means that the system is not tailored to any particular speaker's speech patterns or lip movements, but instead, it is designed to recognize a wide range of speakers. In contrast, general DL-based VSR methodologies are often trained on a specific speaker or a limited set of speakers, making them less effective in recognizing speech from other speakers. Speaker-independent DL-based VSR systems are advantageous because they can be used for a variety of applications that involve recognizing speech from multiple speakers, such as in large group meetings or noisy environments. They are also useful for applications that require quick and accurate recognition of spoken words or phrases, such as in security systems or voice-controlled devices. Moreover, speaker-independent DL-based VSR systems can utilize transfer learning, which uses a pre-trained model to improve predictions on a new but similar problem. This can reduce the required training data and improve the system's performance, making it more efficient than traditional DL-based methodologies that require a large amount of data for training.

\subsection{Futuristic Applications of Speaker Independent VSR}
As mentioned in the above section, VSR technology is particularly useful in situations where traditional audio-based speech recognition systems may not work as effectively. For example, in noisy environments or when a user speaks a language the system may not be familiar with. By using visual data, such as facial expressions and lip movements, VSR can accurately recognize speech and enable more efficient and effective communication between humans and machines. As the demand for more advanced and sophisticated \textbf{Human Machine Interaction (HMI)} systems grows, the potential applications for VSR technology are vast. Some of the futuristic applications of VSR include biometrics \cite{fox2007robust}. This technology is called \textbf{lip biometrics} or \textbf{visual speech biometrics} \cite{chowdhury2022lip, choras2010lip}. Lip biometrics captures and analyzes a video of a person speaking to extract the lip movements and facial features. These features are then used to create a template representing the unique characteristics of the individual's lips and face. This template can be used to compare against other templates in a database to identify a person. One advantage of lip biometrics is that it can be combined with other biometric modalities, such as voice recognition or facial recognition, to create a more accurate and secure system. It is also useful in situations where audio-based biometrics may not be effective, such as in noisy environments or when a person is unable to speak. 

Besides biometrics, speaker independent VSR has promising applications in \textbf{assistive technologies} \cite{kumar2022deep} for individuals with speech impairments. By accurately interpreting visual speech cues, VSR systems can assist in converting lip movements into text or synthesized speech. This can enable individuals with speech difficulties to communicate more effectively, enhancing their independence and quality of life. VSR-based assistive technologies can also find applications in real-time transcription services, aiding individuals in various professional settings, such as meetings, conferences, and classrooms. Also, Speaker Independent VSR can play a crucial role in \textbf{multimedia content analysis} \cite{wang2000multimedia}, particularly in video analysis and understanding. By accurately extracting speech information from visual data, VSR can facilitate tasks such as automatic speech recognition in videos, content indexing, and semantic analysis. This can have applications in various domains, including video search and retrieval, video summarization, video captioning, and content recommendation systems. VSR technology can enhance the efficiency and effectiveness of multimedia analysis, enabling more advanced content understanding and retrieval. These futuristic applications of Speaker Independent VSR demonstrate its potential to revolutionize various domains, including security, assistive technologies, human-computer interfaces, multimedia content analysis, and virtual/augmented reality. As research and development in VSR continue to advance, we can expect further innovations and applications, contributing to the advancement of human-machine communication and interaction.

\subsection{Methodology of Speaker Independent VSR Systems}

\begin{figure}[htbp]
    \centering
    \includegraphics[width = \linewidth]{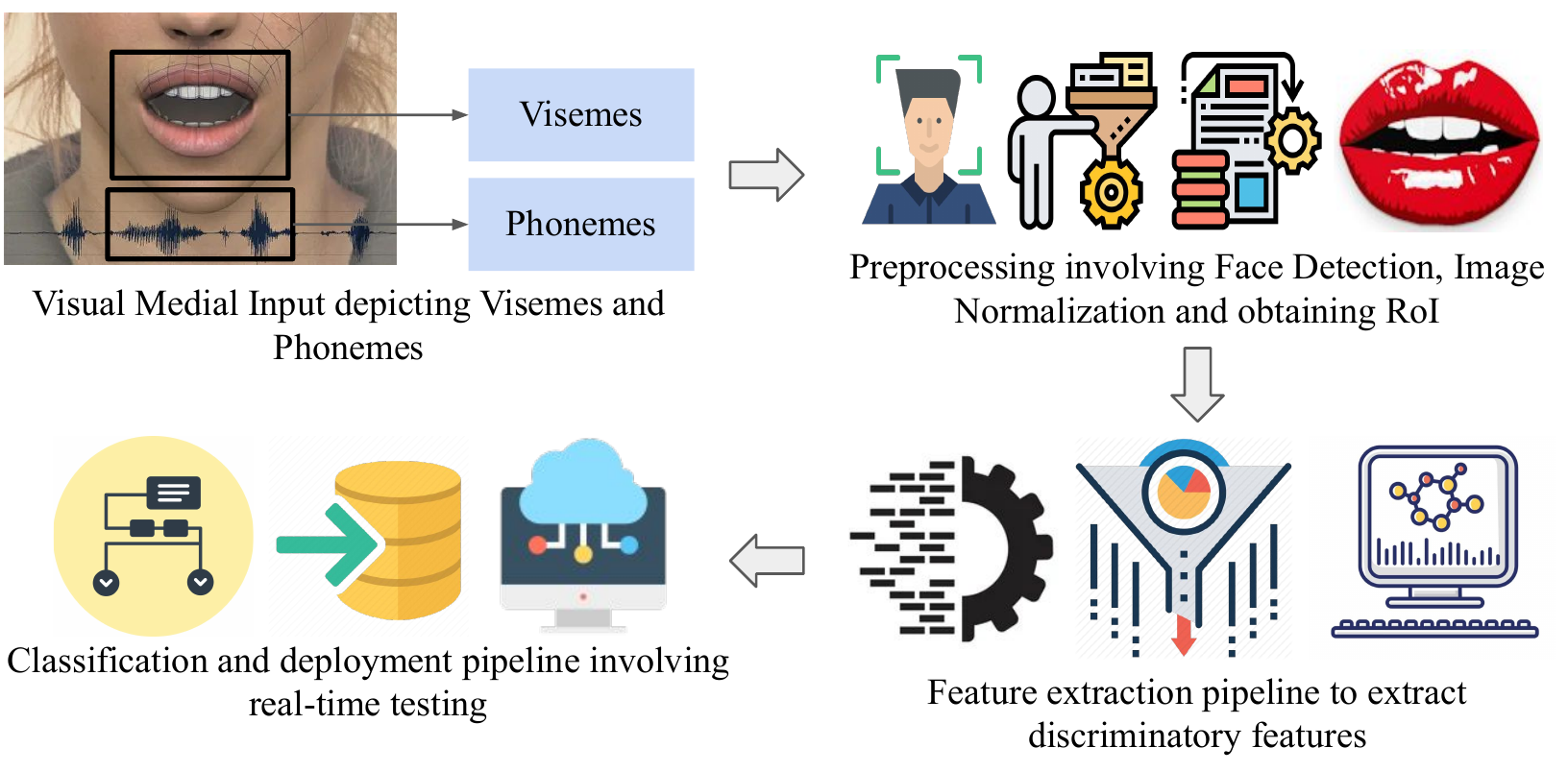}
    \caption{Conceptual overview of VSR}
    \label{fig:CO}
\end{figure}

Given the wide range of potential applications for Speaker Independent VSR systems, it is crucial to clearly understand the underlying processes and concepts involved in the technology. Due to the fact that many words and syllables are pronounced similarly, manual Visual Speech Recognition (VSR) has a low success rate, estimated at only 10\% \cite{altieri2011some}. This highlights the urgent need for more sophisticated procedures to improve the accuracy of VSR systems. The process of the VSR system can be divided into four distinct sub-components: (1) \textbf{Sensory Visual Media Capture}, (2) \textbf{Preprocessing}, (3) \textbf{Feature Extraction and Normalization}, and (4) \textbf{Classification}. The above methodology can be depicted in Fig. \ref{fig:CO}. 

\begin{figure*}[htbp]
    \centering
    \includegraphics[width = \linewidth]{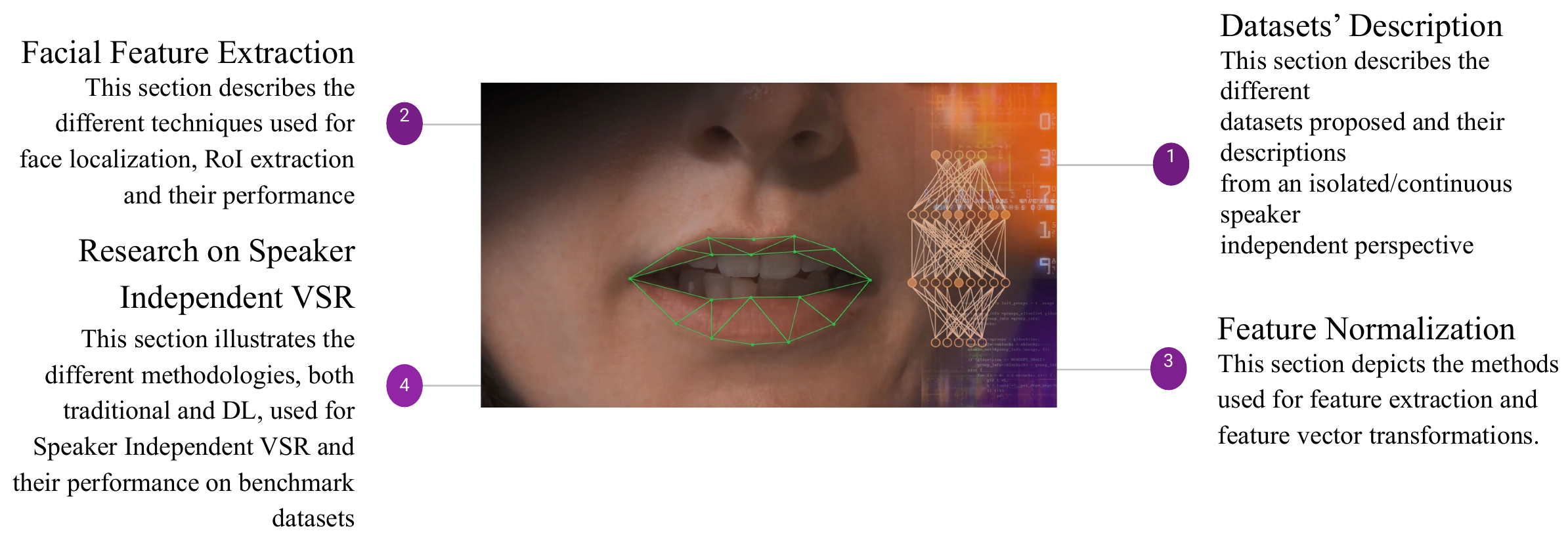}
    \caption{Methodology of the Survey}
    \label{fig:Methodology}
\end{figure*}

\begin{enumerate}
    \item \textbf{Visual media input} is a crucial aspect of speaker independent VSR systems. It involves the use of video recordings of the user speaking a word or sentence, which can be segmented into words, digits, or phrases. These video recordings are also called as \textbf{utterances}. In VSR systems, an utterance refers to a spoken word or a phrase that a speaker utters. It is a sequence of acoustic signals that are processed to recognize the intended speech.  These recordings can be classified into two distinct categories; namely, \textbf{Visemes} \cite{cappelletta2011viseme, bear2016decoding}, and \textbf{Phonemes} \cite{trager1941syllabic, mines1978frequency}. Visemes are the basic visual units of speech that are produced by the movements and positions of the lips, tongue, and other facial features. Visemes are similar to phonemes, the basic units of sound in speech but represent the visual aspects of speech. Visemes are groups of phonemes that are visually indistinguishable from one another, meaning that they look the same when spoken. In layman's terms, visemes simply are lip movements, while phonemes are the sounds from those movements. Typically, several different visemes correspond to each phoneme since different phonemes can be produced with similar visual movements and positions. The exact number and definition of visemes may vary depending on the specific VSR system or application. To develop VSR systems, researchers need to analyze the specific characteristics of lip movements, such as the shape and position of the lips, tongue, and other facial features \cite{pantic2004facial, jiang2002relationship}. This categorization of lip movements helps in identifying which viseme or phoneme is being produced in the video recording, enabling the system to understand spoken language without relying on audio signals.

    \item \textbf{Preprocessing:}  is a critical step in speaker independent VSR, as it helps to improve the accuracy and efficiency of the speech recognition system. The preprocessing pipeline involves several steps, including the \textbf{splitting of visemes into frames, proper frame selection, cropping of face and lip regions of interest  \cite{zhang2022lip, saha2015approach, wang2008roi}, quality enhancement, and filtering}. In the present scenario, this preprocessing step can be considered a fundamental and standard practice for various speaker independent VSR methodologies. The use of traditional computer vision techniques, and in some cases DL, can enable the successful execution of this preprocessing step to enhance the accuracy and reliability of the VSR system. However, the techniques employed may vary depending on the application and other factors. Overall, the preprocessing stage is an essential component of most speaker-independent VSR pipelines in use today.

    \item \textbf{Feature Extraction and Normalization: }After the visual media preprocessing step, feature extraction and normalization are used to extract discriminatory features from the region of interest (RoI) of segmented lip area \cite{schadt2001feature, wang2021multi, xu2021novel}. This step involves mapping the high-dimensional picture data into a lower-dimensional representation and identifying valuable and pertinent information from redundant features. This can be performed using a variety of techniques, including traditional computer vision techniques such as image processing and feature engineering, as well as DL techniques such as CNNs and RNNs. These techniques can automatically extract relevant features from the visual data without the need for explicit feature engineering In speaker-independent VSR, feature extraction is particularly important because it is necessary to recognize speech accurately even when the system has not been trained on the specific speaker's visemes. This requires identifying features that are common across different speakers rather than relying on speaker-specific features. 

    \item \textbf{Classification: }Once feature extraction and normalization are completed, the next step is \textbf{Classification}. In this step, the system reduces the dimensions of the extracted feature vector to predict the class and if possible, the characteristics of the input visual media. For instance, if the dataset includes speakers pronouncing 10 different words, then the number of classes in the system would be 10.

\end{enumerate}

\subsection{Highlights}

The highlights of this paper can be stated as follows:

\begin{enumerate}
    \setlength\itemsep{0.05em}
    \item This review article offers exclusive perspectives and a comprehensive comparison of methodologies employed for feature extraction, normalization, and speech classification.
    \item This article provides a detailed and all-encompassing guide on the various datasets utilized for speaker-independent VSR from the early stages of research in the 1990s until now. 
    \item In this review article, an extensive analysis of conventional algorithms and DL structures for speaker-independent VSR is presented. The comparison between these approaches is conducted based on factors such as accuracy, testing, feature representation, and deployment, providing a comprehensive survey of the field.  
\end{enumerate}

The organization of the paper is as follows: Section \ref{datasets} provides a detailed account of the various datasets employed for VSR. Section \ref{FE} elucidates the different techniques applied for feature extraction and normalization, while Section \ref{FN} outlines the information concerning the diverse architectures employed for prediction and classification. Fig.~\ref{fig:Methodology} illustrates the methodology employed in this survey.  

\section{Overview of the Different Datasets Used in VSR}
\label{datasets}
To perform the task of VSR, a well-defined dataset is deemed essential. Since the task of VSR has many categories like sentence-level VSR and word-level VSR based, Isolated and Continuous, and speaker dependent and speaker independent, it is considered highly essential to choose the appropriate dataset to work upon. When selecting datasets for speaker independent VSR, it is important to consider several criteria. One key factor is the \textbf{size of the dataset}, as having a large amount of data can help improve the system's accuracy. Additionally, the dataset should be \textbf{diverse}, including a wide range of speakers with different accents, dialects, and speech styles. This diversity helps ensure that the system can accurately recognize speech from various sources. Finally, \textbf{proper annotation} of the dataset is also crucial, as it allows for the training and evaluation of the VSR system. By considering these criteria when selecting datasets, researchers can ensure that their VSR system is robust and effective in real-world applications. In this section, we provide a detailed description of the different datasets for the task of VSR along with their comparison as depicted in Table~\ref{Tab:Dataset}. During the 1990s, limited studies focused on developing and evaluating speaker independent VSR systems. As a result, most experiments were conducted using customized datasets, which were not validated on benchmark datasets. This limited evaluation of speaker independent VSR systems was due to several factors, including the lack of standard datasets, limited computational resources, and the complexity of developing robust and accurate VSR systems. However, some studies did evaluate both speaker-dependent and speaker-independent VSR systems on a few datasets, including the Tulips1 and M2VTS datasets. \textbf{Tulips1} \cite{movellan1996channel} is a dataset for audio-visual speech recognition. This dataset consists of 12 speakers, who are undergraduate students of UCSD who are native speakers, each uttering the first four digits of English, and each speaker repeats the digits twice. The audio portion is captured at 11127 Hz with 8 bits per sample. The video portion consists of 934 grayscale lip frames with a size of 100 × 75 pixels, captured at a rate of 30 fps on a frontal pose. Another popular dataset used in the 90s, the \textbf{M2VTS dataset} \cite{kittler1997combining}, is a publicly available benchmark dataset for speaker verification and identification research. It was created in the late 1990s by the University of Surrey and the Swiss Federal Institute of Technology in Lausanne, Switzerland. The dataset consists of video and audio recordings of 37 speakers, 22 of whom are male and 15 of whom are female, all of them being native Swiss speakers. Each speaker was recorded multiple times, resulting in a total of 1,000 video and audio recordings collected over a period of 4 months. The recordings were captured using a fixed camera and a close-up view of the speaker's face, allowing for the analysis of lip movements and facial expressions during speech. 

\begin{figure}[htbp]
    \centering
    \includegraphics[width = \columnwidth]{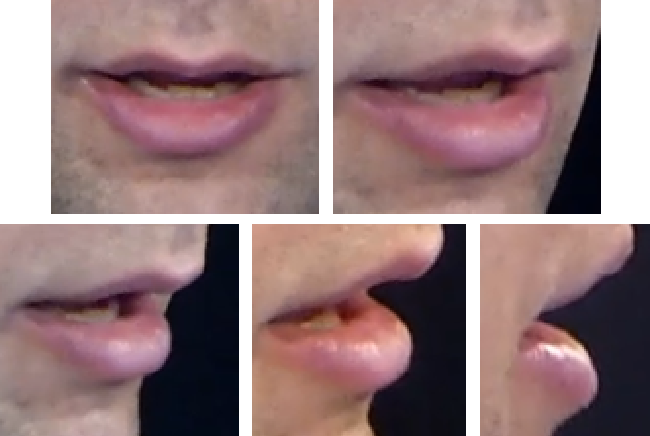}
    \caption{Frames from the OuluVS2 dataset}
    \label{fig:Olu}
\end{figure}

The \textbf{AVLetters} \cite{982900} dataset consists of 10 speakers (five men and five women), each uttering the letters {\it A-Z} 3 times, so a total of 780 utterances is one of the most used alphabet recognition datasets. It consists of native english speakers and is recorded at the University of East Anglia. The audio portion is recorded using 26 MFCC (Mel-frequency Cepstral Coefficient), and the video portion is recorded with a size of 376 $\times$ 288 pixels. The dataset is divided into two categories: training (first two utterances of every letter of every speaker); and testing data (third utterance from all speakers). \textbf{AVLetters2} \cite{inproceedings} was recorded at the University of Surrey in 2008, a revised version of the AVLetters dataset. This dataset consists of 5 speakers, each uttering the letters {\it A-Z} 7 times, so a total of 35 videos per letter. High Definition cameras with a resolution of 1920 $\times$ 1080 were used to capture videos of this dataset in RGB colorspace. This dataset is superior to its predecessor version as it contains more classes with enhanced resolution of the recordings. Another dataset, \textbf{OuluVS} \cite{5208233}, is a dataset containing 20 speakers, each uttering 10 sentences and every speaker uttering each sentence five times. All image sequences of the OuluVS dataset are segmented, and the mouth area is determined by the manually marked eye position in every frame. Every image in all the sequences has a resolution of 720 $\times$ 576 pixels and is captured at  25 fps. Unlike the previous datasets mentioned above, this dataset deals with sentences in an isolated framework. An enhanced version of this dataset, also called the OuluVS2 dataset, contains multidimensional images as depicted in Fig. \ref{fig:Olu}. Applying the concept of multidimensionality, a dataset named \textbf{Lip Reading in the Wild (LRW)} is proposed. LRW is an audio-visual dataset comprising 500 distinct classes of words spoken by more than 1000 people, both native and non-native speakers. This dataset is made up of 1.16-second-long MPEG4 clips that were each captured at 25 fps. The dataset is divided into training sets (800 utterances from each class) and validation and test sets containing 50 utterances from each class.

\begin{figure}[htbp]
    \centering
    \includegraphics[width = \columnwidth]{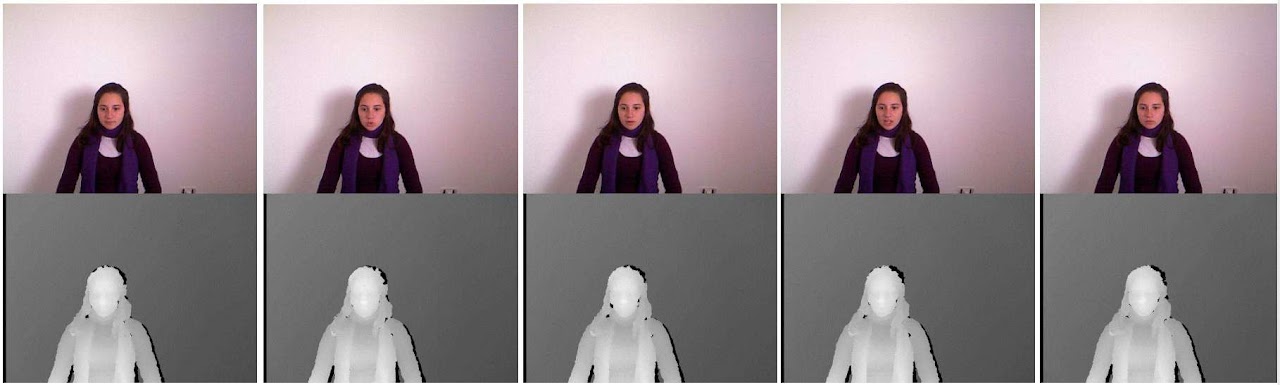}
    \caption{Frames from the MIRACL-VC1 dataset}
    \label{fig:MC}
\end{figure}

Since the proposal of the LRW dataset, several other datasets involving the multidimensional data concept have also been proposed. One such example of a dataset is the \textbf{Lip Reading Sentences 2 (LRS2)} dataset \cite{Chung17}. Most videos in the LRS2 dataset come from BBC news and TV shows. The length of each sentence is up to 100 characters. According to the broadcast date, the dataset is divided into 3 sets: training, validation, and test set, to avoid similarity in the data. It is complex since it has a wide range of head poses and thousands of speakers without speaker labels. All the clips are captured with a resolution of 160 $\times$ 160 pixels at 25fps. Another dataset, the \textbf{LRS3-TED} \cite{DBLP:journals/corr/abs-1809-00496}, is a dataset for both visual and audio-VSR. This dataset comprises 5594 English TED and TEDx talks gathered from YouTube, and it consists of more than 400 hours of video. The trimmed face recordings are provided as a .mp4 file, encoded with the h264 codec with a resolution of 224 $\times$ 224 pixels at 25 fps. The dataset is divided into pre-train, train-Val, and test. A dataset involving the pronunciation of digits in a multidimensional space is the \textbf{AVDigits} dataset \cite{7780758}. It consists of 53 speakers, each uttering the 0–9 digits five times each, and 39 speakers, each uttering 10 sentences five times each. The videos are captured in 3 different positions (0, 45, \& 90), and three different speech modes (standard, whisper, \& silent), and the speakers were asked to close their mouths at the start and end of uttering the digits. The video portion of each utterance is captured with a resolution of 1280 $\times$ 780 pixels at 30 fps. One of the significant datasets for isolated sentence and word-level VSR is the \textbf{MIRACL-VC1} dataset \cite{rekik2014new}. The major highlight of this dataset is that it also contains depth images of the speakers, as illustrated in Fig. \ref{fig:MC}, which adds to the training of any DL model. The dataset consists of 10 words and 10 phrases, with 15 speakers (5 men and 10 women) each uttering the words and phrases ten times each, totaling 3000 utterances. Every utterance in the sequence is an image of size 640 x 480 pixels with 25fps. 

\begin{table*}[htbp]
\label{Tab: datasets}
	\centering
	\caption{Comparison between the existing datasets for isolated VSR}
	\label{Tab:Dataset}
	\resizebox{\linewidth}{!}{\begin{tabular}{|c|c|c|c|c|c|c|c|c|c|c|c|c|}
	\hline
	\textbf{Dataset} & \textbf{Language }& \textbf{Year} & \textbf{I/C} & \textbf{Segment} & \textbf{Speakers} & \textbf{Classes} & \textbf{Utterances} & \textbf{Resolution} & \textbf{Pose} & \textbf{Avg Sample Size} \\ 
\hline

Tulips1 \cite{movellan1996channel} & English & 1996 & Isolated & Digits & 12 & 4 & 96 & 100 $\times$ 75 & Frontal & - \\ \hline 

M2VTS \cite{kittler1997combining} & English & 1997 & Isolated & Digits & 37 & 10 & 1000 & 768 $\times$ 576 & Frontal & 1.5s \\ \hline

AVLetters \cite{982900} & English & 1998 & Isolated & Alphabets & 10 & 26 & 780 & 376$\times$ 288 & Frontal & -\\
\hline
AVLetters2 \cite{inproceedings} & English & 2008 & Isolated & Alphabets & 5 & 29 & 910 & 1920$\times$ 1080 & Frontal & - \\
\hline
MIRACL-VC1 \cite{rekik2014new} & English & 2011 & Isolated & Words & 15 & 10 & 1500 & 640 $\times$ 480 & Frontal & 1s \\
\hline

OuluVS \cite{5208233} & English & 2015 & Isolated & Sentences & 20 & 10 & 1000 & 720$\times$ 576 & Frontal & 3s \\
\hline
LRW \cite{Chung16} & English & 2016 & Continuous & Words & $>$ 1000 & 500 & 400000 & 256$\times$ 256 & -30 $\sim$ 30 & 3s \\
\hline
LRS2 \cite{Chung17} & English & 2017 & Isolated & Sentences & $>$ 1000 & 17428 & 118116 & 160$\times$ 160 & -30 $\sim$  30 & 3.75s \\
\hline
LRS3-TED \cite{DBLP:journals/corr/abs-1809-00496} & English & 2018 & Isolated & Sentences & $>$ 1000 & 70000 & 165000 & 224$\times$ 224 &  -90 $\sim$  90 & 3.2s\\
\hline
AVDigits \cite{7780758} & English & 2018 & Isolated & Digits & 53 & 10 & 795 & 1280$\times$ 780 & 0,45,90 & - \\
\hline
AVDigits \cite{7780758} & English & 2018 & Isolated & Sentences & 39  & 10 & 5850 & 1280$\times$ 780 & 0,45,90 & -\\
\hline
CAS-VSR-W1k \cite{tian2022lipreading} & Mandarin & 2019 & Continuous & Word & 2000 & 1000 & 718,018  & Naturally distributed & Natural & - \\
\hline
CMLR \cite{zhao2019cascade} & Mandarin & 2021 & Isolated & Sentences & 11 & 9 & 102076  & 64 $\times$ 128 & Frontal & - \\
\hline

GLips \cite{schwiebert2022multimodal} & German & 2022 & Continuous & Words & 100 & 500 & 250000 & 256 $\times$ 256 & Frontal & - \\
\hline
CN-CVS/Speech \cite{chencn} & Mandarin & 2022 & Continuous & Sentences &  2529 &  $\sim$ 75 & 193,329  & 640 $\times$ 480 & Natural & - \\
\hline
CN-CVS/News \cite{chencn} & Mandarin & 2022 & Continuous & Sentences & 28 & $\sim$ 465 & 13016 & 640 $\times$ 480 & Frontal & - \\
\hline

RUSAVIC \cite{ivanko-etal-2022-rusavic} & Russian & 2022 & Isolated & Sentences & 20 & 72 & 14000  & 1920 $\times$ 1080 & -30 $\sim$ 30 &  - \\
\hline
OLKAVS \cite{park2023olkavs} & Korean & 2023 & Isloated & Sentences & 1107 & $>$ 100 &  250000 & 1920$\times$ 1080 & 0,45,90 & - \\
\hline
	\end{tabular}}
\end{table*}

As researchers aim to develop VSR systems that can recognize speech in languages other than English, non-English datasets have gained importance. These datasets are chosen based on criteria such as dataset size, diversity, and proper annotation. \textbf{OLKAVS} \cite{park2023olkavs}, for instance, is a Korean dataset comprising over 1150 hours of audiovisual data, while \textbf{Glips} \cite{schwiebert2022multimodal} includes over 250000 utterances of German speakers. \textbf{CAS-VSR-W1k} \cite{tian2022lipreading} is a Mandarin dataset that consists of over 1,000 sentences spoken by 2000 speakers, and \textbf{CN-CVS} \cite{chencn} includes over 122.6 hours of Chinese audiovisual data. \textbf{RUSAVIC} \cite{ivanko-etal-2022-rusavic} is a Russian dataset that includes over 14000 utterances of Russian audiovisual data, while \textbf{CMLR} \cite{zhao2019cascade} includes over 102076 video clips of Mandarin speakers. These datasets differ in various aspects such as the number of speakers, the duration of recordings, and the number of video clips. These factors can affect the performance and robustness of VSR models trained on them. Despite the variations, these datasets provide valuable resources for developing and evaluating VSR models for a range of languages. These datasets are open-source with the exception of M2VTS and OuluVS having been replaced by their upgraded versions, which enables researchers worldwide to access and use them. In the subsequent sections, we illustrate the different methodologies applied to these datasets for preprocessing, feature extraction, and prediction. This would give readers an insight into the techniques used in the field and the potential of these datasets in advancing the performance of VSR models.

\section{Feature Extraction}
\label{FE}

\begin{figure}[htbp]
    \centering
    \includegraphics[width = \columnwidth]{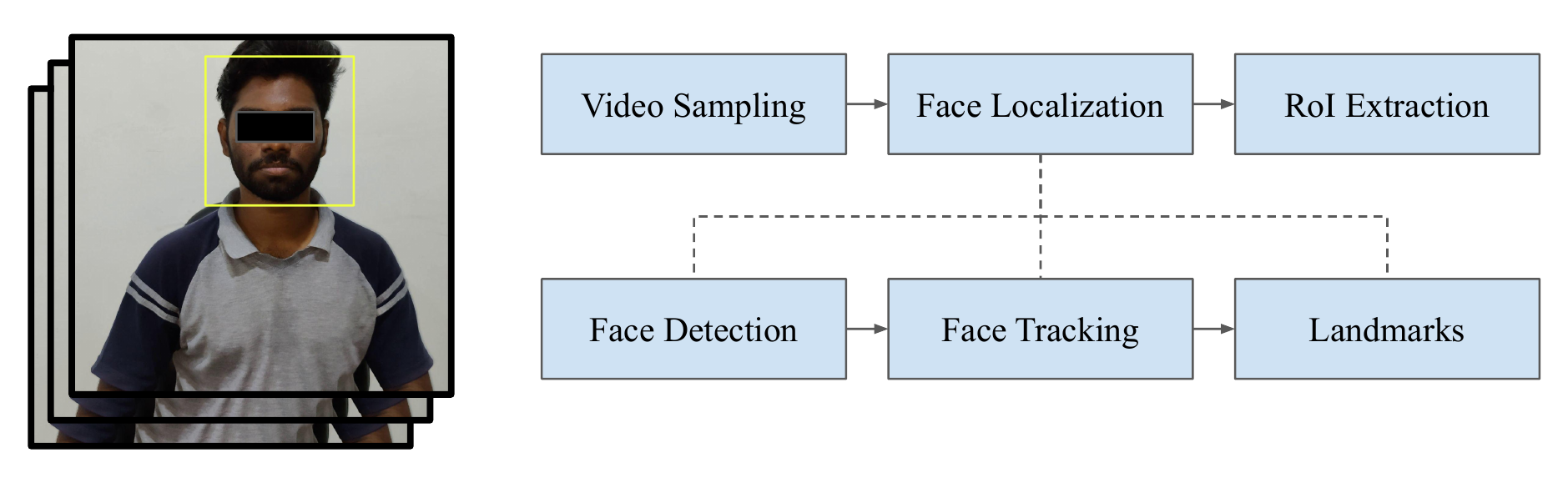}
    \caption{Preprocessing Pipeline}
    \label{fig:PP}
\end{figure}

\textbf{Preprocessing} and \textbf{feature extraction} are essential steps in the process of speaker independent VSR. Preprocessing aims to enhance the quality of the data by eliminating unwanted distortions or enhancing specific visual properties necessary for subsequent processing and analysis tasks. Figure \ref{fig:PP} depicts the face localization and preprocessing process used for extracting the lip region of interest. This section outlines the various techniques used for dataset preprocessing and lip region of interest extraction.

The concept of \textbf{fuzzy clustering}, which allows a data point to belong to more than one cluster, has become increasingly important in many DL classification tasks \cite{yang1993survey, miyamoto2008algorithms, xie1991validity}. In the context of VSR, a novel technique for lip contour extraction was proposed by \textbf{Srinivasa Rao Chalamala et al. \cite{7066486}}, which employs fuzzy clustering with \textbf{active contour modeling (ACM)} \cite{chan1999active} and elliptic shape information. The proposed approach involves the extraction of lip structure using the fuzzy membership distribution, followed by the application of an ellipse function to refine the extracted lip structure. The authors defined their base model as an ellipse enclosing the lip area, which was then used to develop active contours to precisely match the lip's contour. By combining image and model-based lip contour extractions, the proposed technique aims to increase the accuracy of lip contour extraction for VSR tasks. Overall, this research presents a promising approach to improve lip contour extraction accuracy for VSR, utilizing the concepts of fuzzy clustering and active contour modeling. The use of an ellipse function to refine the lip structure is a novel addition to existing techniques, and the combination of image and model-based extractions is expected to provide more robust and accurate results. However, it is noteworthy that the authors do not explicitly address the effectiveness of their proposed approach across different populations. Specifically, they do not provide explicit insights into whether the method performs consistently well across individuals of different ages, ethnicities, or other demographic factors. \textbf{The researchers contend that the ACM, when applied solely by itself, fails to yield accurate lip contours}. They specifically highlight scenarios in which the pixel values in the lip region and its surrounding areas lack distinct separation, making it challenging for the ACM to accurately delineate the lip boundaries. This aspect leaves room for further investigation and validation of the approach's generalizability and applicability to diverse populations. 

\textbf{Chengjia Yang et al. \cite{6933639}} proposed an improved version of the region growing algorithm \cite{mehnert1997improved} which  utilizes the RGB colorspace for lip contour extraction. The proposed method estimates the association between the color form and adjacent pixels, thereby automatically selecting seed pixels. Subsequently, the \textbf{watershed algorithm} is employed after a series of image processing operations to extract the lip contour region. The experimental analysis conducted on the proposed approach demonstrates its straightforward nature and effectiveness in extracting the lip contour region, achieving an accuracy rate of 80\%. Notably, the approach leverages the full RGB color space, eliminating the necessity for color space conversion, which in turn saves both \textbf{memory space} and \textbf{computing time}. \textbf{Given the absence of comprehensive details regarding the customized dataset utilized in the study, it becomes challenging to ascertain the generalizability of the methodology across different ethnicities}. This lack of specificity pertaining to the dataset poses limitations on the broader applicability and effectiveness of the proposed approach across diverse ethnic populations. \textbf{Sukesh Kumar Das et al. \cite{8487538}} proposed an automated mechanism for lip contour detection, utilizing segmentation based on pixel characteristics and piecewise polynomial interpolation. The initial RGB input color frame undergoes binary image thresholding. The region of interest (ROI) is identified by constructing a mathematical model that divides the data into two distinct clusters, namely, lip pixels and non-lip pixels, with discriminatory color information. The experimental analysis indicates that the performance of this proposed methodology is slightly better for females \textbf{(accuracy = 99.46\%)} than for males \textbf{(accuracy = 98.84\%)} due to the greater color discriminant intensity weights between female subjects' lips and non-lip areas. \textbf{Although it is important to acknowledge the limitations of solely relying on the performance on a single dataset, the successful results obtained on the GRID Database provide encouraging evidence of the methodology's efficacy}. As the GRID Database consists of recordings from individuals of various ethnic backgrounds, the methodology's high accuracy on this dataset implies its potential for generalization across different ethnicities.

\textbf{Active contour} is a popular segmentation technique in image processing that employs energy forces and limitations to extract essential pixels from an image for further analysis and processing. The primary purpose of active contours is to define smooth shapes in images and identify asymmetrical shapes. \textbf{Xin Liu et al. \cite{5597771}} proposed a localized active contour model with automatic parameter assignment for lip segmentation. The proposed technique involves identifying the lips' initial edge landmarks to determine the lips' minimum-bounding ellipse. At each iteration, local energies are computed at each point along the curve, resulting in the deformation of the evolving curve. The iterations continue until the minimum bounding ellipse is the best fit. With an impressive \textbf{accuracy rate of 96.4\%}, the authors confidently assert that their methodology successfully extracted lip contours, even in cases involving deformable or irregular lips. The proposed approach exhibits a higher tolerance towards challenges such as uneven illumination, rotation, deformation, and the presence of teeth and tongue, which can significantly impact the accuracy of lip contour extraction. Moreover, the authors conducted a thorough examination of the unsatisfactory results, which accounted for only \textbf{3.6\%} of the cases, and identified that these instances were primarily attributed to either \textbf{poor contrast between the lip and surrounding skin regions or the presence of noticeable beard effects around the lips}. 

In contrast to typical ACMs with global information, \textbf{Chin et al. \cite{6111275}} proposed a region-based ACM with local energy terms, assisted by the watershed algorithm. The framework includes localized energy estimation based on watershed segmentation, delivering a more satisfactory contour extraction mechanism than global energies computation. Additionally, a modified H mechanism is presented to efficiently identify lip feature landmarks, which counteracts the delicate nature of ACM with the initial contour position. The proposed framework is experimented with the \textbf{CUAVE} and \textbf{XM2VTS} benchmark video datasets, and results show a \textbf{Percentage of Overlap (POL) of 90\% and 86.2\%}, respectively. 
The authors assert that their proposed system surpasses existing methodologies when confronted with scenarios where the color differences between the lips and surrounding skin regions are indistinct. \textbf{The localized analysis employed in their approach, which focuses on local details rather than solely seeking global energies, proves to be more effective in these challenging situations}. Additionally, the authors claim that their algorithm demonstrates superior lip contour detection for subjects with darker skin tones and individuals with facial hair, specifically mentioning black subjects. Since the experimentation encompassed two benchmark datasets, \textbf{it can be reasonably concluded that the proposed algorithm exhibits proper generalization across different ethnicities}. 

\textbf{Alan Wee-Chung Liew et al.} \cite{1220299} proposed a spatial fuzzy clustering algorithm \cite{guo2015new} for binary-class clustering and segmentation tasks. The algorithm leverages feature space data and the spatial associations between adjacent pixels to achieve successful segmentation. This algorithm includes appropriate preprocessing and postprocessing phases to enhance the image quality further. According to the results obtained, the proposed algorithm demonstrated successful segmentation, achieving a \textbf{POL of 95\%} or a \textbf{Segmentation Error (SE) of 5\%.} However, it is important to note that a portion of the error can be attributed to the mismatch between the parametric lip model employed and the actual lip characteristics. Based on the author's experience, this type of error typically accounts for approximately 2\% to 5\% of both POL and SE, with the specific percentage varying depending on the actual shape of the lips. \textbf{It should be noted that the algorithm does not incorporate the consideration of facial hair, such as mustaches and beards}. \textbf{Shu-Hung Leung et al. \cite{1262013}} introduced a spatial-color-based fuzzy clustering framework for lip region segmentation. This framework called \textbf{fuzzy c-means with shape function (FCMS)}, utilizes color information and spatial distance to extract the lip region. The proposed framework presents a new dissimilarity measure incorporating the shape function in its embeddings, which improves the segmentation performance. The framework's implementation involves color transformation, image preprocessing, lip extraction with FCMS, and postprocessing. However, the computational time complexity of the proposed FCMS is nearly \textbf{1.78 times} that of the classical fuzzy c-means (FCM) algorithm. In conclusion, the research papers by Alan Wee-Chung Liew et al. \cite{1220299} and Shu-Hung Leung et al. \cite{1262013} introduce novel approaches for image segmentation using spatial fuzzy clustering algorithms. These techniques are valuable in extracting essential features from an image by leveraging spatial and color information. The proposed frameworks demonstrate improved segmentation performance, \textbf{but the computational complexity should be considered when applying them in real-world applications}. 

\textbf{Yao Wen Juan et al. \cite{5579830}} proposed a real-time lip localization framework comprising two phases: lip region location and lip tracking. Firstly, the framework detects the face and eyes using \textbf{AdaBoost} \cite{schapire2013explaining, hastie2009multi} and \textbf{Haar Features} \cite{whitehill2006haar, wilson2006facial} and, based on their association with the mouth, locates the lips. Subsequently, the lips are tracked using the $\mathit{a}$ component of the Lab color space. In the study, it is reported that the proposed lip tracking method achieved accurate tracking for \textbf{98\%} of the lips. The authors substantiate this by demonstrating that the $a_{component}$ of the lab color space exhibits superior separation ability compared to other color space components. Additionally, the authors highlight the successful improvement of their lip location method in enhancing lip tracking performance across \textbf{various lighting conditions, diverse lip shapes, and different head poses}. In a similar vein, a color-based geometrical algorithm for lip region extraction was proposed by \textbf{Shemshaki et al. \cite{6121606}}. This algorithm defines geometrical rules for skin and lip distributions in the \textbf{chromatic and YCbCr} color spaces, respectively, to extract pixels containing skin and lip separately in an image. After segmenting the lip region, the proposed algorithm further improves its accuracy by performing perpendicular and horizontal accumulation curves, achieving a detection rate of 93.33\%. The results obtained from the study provide evidence that the proposed method is capable of accurately detecting the lip region in images with complex backgrounds. The authors also claim that the method \textbf{demonstrates efficiency in terms of both implementation and execution time, making it a practical solution}. Additionally, the proposed method exhibits high precision in detecting lips in images with rotated faces. It should be noted that the study does not provide specific information regarding the performance of the methodology on different ethnicities. The authors do not explicitly address the methodology's effectiveness or generalizability across diverse ethnic populations.

\begin{table*}[t]
	\centering
	\caption{Comparison between existing Feature Extraction works}
	\label{Related:feature}
	\resizebox{\linewidth}{!}{
	\begin{tabular}{|c|p{46mm}|p{38mm}|c|}
	
	\hline
	\textbf{Existing Works}& 
	\centering \textbf{Methodology}& 
	\centering \textbf{Dataset}& 
	\centering \textbf{Feature Details}
	\tabularnewline
	\hline

    Srinivasa et al. \cite{7066486} &
	 \centering Fuzzy clustering  + active contour modelling &
	 \centering VidTimit database &
	 \centering Lip Contours 
	 \tabularnewline
	 \hline
  
   Chengjia Yang et al. \cite{6933639} &
	 \centering RGB–based improved region growing &
	 \centering Custom lip images&
	 \centering Lip Contours
	 \tabularnewline
	 \hline
  
    Sukesh et al. \cite{8487538} &
	 \centering Pixel-based segmentation and piece-wise polynomial fitting &
	 \centering GRID database &
	 \centering Lip Contours
	 \tabularnewline
	 \hline
     Xin Liu et al. \cite{5597771} &
	 \centering Localized Active Contour Model with Automatic Parameter Selection &
	 \centering CVL  and GTAV face database  &
	 \centering Lip Contours
	 \tabularnewline
	 \hline
    Siew et al. \cite{6111275} &
	 \centering Watershed-based active contour model and modified $H_{\infty}$ &
	 \centering CUAVE, XM2VTS  database  &
	 \centering Lip contour detection and tracking 
	 \tabularnewline
	 \hline
  
	 Alan et al. \cite{1220299} &
	 \centering Spatial Fuzzy Clustering &
	 \centering Homebrew, XM2VTS, AR face &
	 \centering Lip Segmentation
	 \tabularnewline
	 \hline
	 
	 Shu-Hung et al. \cite{1262013} &
	 \centering Fuzzy Clustering + Elliptic Shape Function &
	 \centering Custom database &
	 \centering Lip Segmentation
	 \tabularnewline
	 \hline

  Yao WenJuan et al. \cite{5579830} &
	 \centering OpenCV and $a_{component}$ of Lab color space& 
	 \centering CR face and dual-mode video database  & 
	 \centering Lip Localization and Tracking
	 \tabularnewline
	 \hline

	 Mehrdad et al. \cite{6121606} &
  \centering Geometrical Model Of Color Distribution& 
  \centering Bao Image database & 
  \centering Lip Segmentation
	 \tabularnewline
	 \hline
  
   MA Xinjun et al. \cite{7554449} &
	 \centering Improved Jumping-Snake Model &
	 \centering Custom Database   &
	 \centering Lip point’s detection 
	 \tabularnewline
	 \hline

  Zhiyong et al. \cite{4730358} &
      \centering Dynamic deformable templates &
      \centering CU-TTAVS Database  &
      \centering Lip Tracking
	 \tabularnewline
	 \hline
	 
	 Parisa et al. \cite{5495373} &
	 \centering Fuzzy logic approach with particle filter &
    \centering Custom Database  &
	 \centering Lip Tracking
	 \tabularnewline
	 \hline

	 Behrooz et al. \cite{6599705} &
	 \centering Modified Lip-Map Algorithm&
	 \centering CVL, IMM, GTAV, and CID Databases  &
	 \centering Lip detection 
	 \tabularnewline
	 \hline

  Ma et al. \cite{ma2022training} &
  \centering RetinaFace + FAN &
  \centering LRW Dataset &
  \centering Lip Localization and Tracking 
  \tabularnewline
  \hline

\end{tabular}}
\end{table*}

\textbf{Xinjun et al. \cite{7554449}} proposed an improved jumping-snake model for lip segmentation that considers the geometric parameters of the lips. This model includes a \textbf{division-detection procedure}, which involves the estimation of the lip region and the detection of lip landmarks. The method starts by utilizing the frontal face template matching technique to segment the lips, which is followed by the application of the improved jumping-snake technique to identify the lip landmarks. The validation of the methodology on a customized dataset consisting of Chinese speakers, with an average \textbf{accuracy of 90\%}, is indeed valuable information. However, it is important to recognize that the \textbf{absence of other ethnic speakers in the dataset limits the ability to directly assess the performance of the methodology on different ethnicities}. Another approach to lip segmentation is the use of dynamic deformable templates, which was proposed by \textbf{Zhiyong Wu et al. \cite{4730358}}. This method utilizes different templates based on varying lighting conditions, head poses, and lip shapes. For variable illumination conditions, a nostril-searching technique is employed for lip segmentation. For variable head poses, the lips are tracked on facial images with the inclusion of a dynamic mechanism for conventional systems. Finally, for variable lip shapes, the lips are tracked based on the contours dynamically. The proposed method was evaluated on the \textbf{CU-TTAVS dataset} and achieved promising results. By incorporating \textbf{"dynamicity"} into the deformable templates, the proposed method addresses the challenges posed by changes in \textbf{head pose, face illumination, and lip shapes}. These variations can significantly impact the accuracy and reliability of lip-tracking algorithms. The dynamic deformable templates, being adaptable and flexible, effectively account for these variations, resulting in improved tracking performance. These approaches demonstrate the advancements in lip segmentation techniques and highlight the importance of considering various factors such as geometric parameters, lighting conditions, head poses, and lip shapes in achieving accurate lip segmentation. Such advances in lip segmentation techniques can have practical applications in various fields, such as speech recognition, facial expression analysis, and biometric identification. 

Improving the accuracy of lip tracking and segmentation remains a challenging task, especially for extremely low-contrast images. To tackle this problem, \textbf{Darvish Zadeh et al. \cite{5495373}} proposed a novel approach that utilizes an improved particle filter lip tracker based on fuzzy logic. The method is suitable for non-linear modeling and is appropriate for dealing with non-Gaussian noise. The proposed technique involves sampling lips using a modified particle filter, and the evaluation of features is accomplished based on fuzzy rules. The results of the proposed particle filter lip tracker indicate that the detection rate is approximately 78\% within a tolerance of 5\% of distances, which is a promising outcome. It is important to note that the methodology under consideration was validated using a \textbf{customized dataset}. While this validation provides valuable insights into the performance of the method on the specific dataset, it does not provide direct information about the methodology's effectiveness on different ethnicities. In addition, \textbf{Behrooz et al. \cite{6599705}} proposed an improved lip-map algorithm for lip region segmentation. In this approach, the lower half part of the image is considered as the input, followed by an improved lip map algorithm, which is further transformed into a wavelet form. The algorithm emphasizes the lip region during these phases, which helps to determine the lip regions and locations accurately. Finally, the lip pixels are extracted from the skin pixels with Top Hat transformation. A significant advantage of this work lies in its evaluation on multiple benchmark datasets, which provides evidence of its ability to generalize across different ethnicities. By evaluating the methodology on various datasets, researchers were able to assess its performance in diverse scenarios and populations, thereby enhancing confidence in its generalizability. \textbf{Furthermore, the reported high accuracies of 99.1\%, 98.0\%, 98.5\%, and 97.8\% on the CVL, IMM, GTAV, and CID databases, respectively, highlight the remarkable efficiency of the methodology.}

In our survey, we recognize the importance of analyzing the feature extraction techniques employed on widely-used datasets, such as the \textbf{LRW dataset}. To address this, we included an overview of the feature extraction works that have been performed on this dataset. We acknowledge the significance of this dataset in the field of VSR and aim to provide a comprehensive analysis of the techniques employed for feature extraction. \textbf{Ma et al.} \cite{ma2022training} utilized the RetinaFace tracker \cite{9589577} to detect the faces present in the video frames. Following this, the authors utilized the \textbf{Face Alignment Network (FAN)} \cite{bulat2017far} for the detection of facial landmarks, which helps to accurately locate the lip area in each frame. They also normalized the frames by removing any size and rotation differences by registering the faces to the mean face in the training set. In order to focus on the mouth region of the face, they cropped a bounding box of size 96 × 96 around the lip area. Furthermore, they applied normalization to each frame by subtracting the mean and dividing by the standard deviation of the training set. This helped them to standardize the data and ensure that the model could effectively learn from the input data without being biased towards any particular distribution. Similar methods of ROI extraction were stated in \cite{koumparoulis2022accurate, 9909819, s23042284}

\textbf{Facial feature extraction} plays a vital role in the success of speaker independent VSR systems, and accurate localization and extraction are critical components of facial recognition pipelines. The section highlights various methods used for facial feature extraction and face localization, with Table \ref{Related:feature} summarizing the methodologies of different techniques. The datasets used in these works vary from VidTimit, GRID, CVL, GTAV, XM2VTS, AR face, CR face, dual-mode video database, Bao Image database, CU-TTAVS, CID, IMM, to LRW. Most of these datasets contain a large number of images or videos, with different angles, poses, and expressions of the subjects. \textbf{Fuzzy clustering} has emerged as a popular choice for lip segmentation due to its ability to accurately classify pixels as part of the lip or non-lip area, even when the lip area is partially obscured or blended with the surrounding skin. Furthermore, fuzzy clustering can handle variations in lighting and noise, making it a practical choice for lip segmentation in real-world environments. Regarding lip contour detection, the majority of works used \textbf{active contour models}, while some utilized the watershed algorithm. Active contour models excel at handling variations in lip texture, shape, and illumination, making them well-suited for lip contour detection in challenging conditions. Moreover, active contour models can adapt to lip shape and motion, making them ideal for tracking lip movements over time. These models can also detect subtle lip movements that are crucial for accurate VSR. It is also essential to understand the computational complexity of It is worth noting that accurate facial feature extraction requires a correct measure to represent the extracted features. This measure can be in the form of feature vectors that represent the unique characteristics of each face. Therefore, it is essential to use appropriate methods for feature extraction to ensure that the feature vectors accurately represent the individual's facial features. In the subsequent section, we will focus on different techniques used for transforming the feature vectors obtained from feature extraction into suitable forms for various VSR architectures. 

\section{Feature Normalization}
\label{FN}
The subsequent step of this pipeline includes feature normalization. The significance of image normalization is that it helps in the faster convergence to find more similarities based on measured feature values from the dataset while training any DL model. In this section, we analyze the different methodologies used to perform the task of image normalization. One of the earliest methodologies of facial detection and normalization proposed by \textbf{Wang et al. \cite{5623095}} uses the similarity of skin color pixels to detect the face in an image. The noise removal is performed by expansion and corrosion after the detection of skin. The methodology uses binarization, image cutting, and enlargement for image normalization. The methodology generalizes well on a diverse set of faces having different patterns of facial and also due to the presence of a simple methodology, it can be considered \textbf{computationally efficient}. However, even due to the presence of diverse facial features, the methodology could face limitations due to the \textbf{absence of evaluation on benchmark datasets}. \textbf{Kim et al. \cite{981881}} proposed a methodology involving the usage of the defined facial section pattern's symmetry and its characteristics for normalizing range data. Surface curvatures can indicate a local surface form and are utilized for accurate feature detection. The significant advantage of this methodology is the inclusion of 3D feature maps for normalization. However, the work faces the same limitation of lack of testing on benchmark datasets. 

\begin{figure}[htbp]
    \centering
    \includegraphics[width = \linewidth]{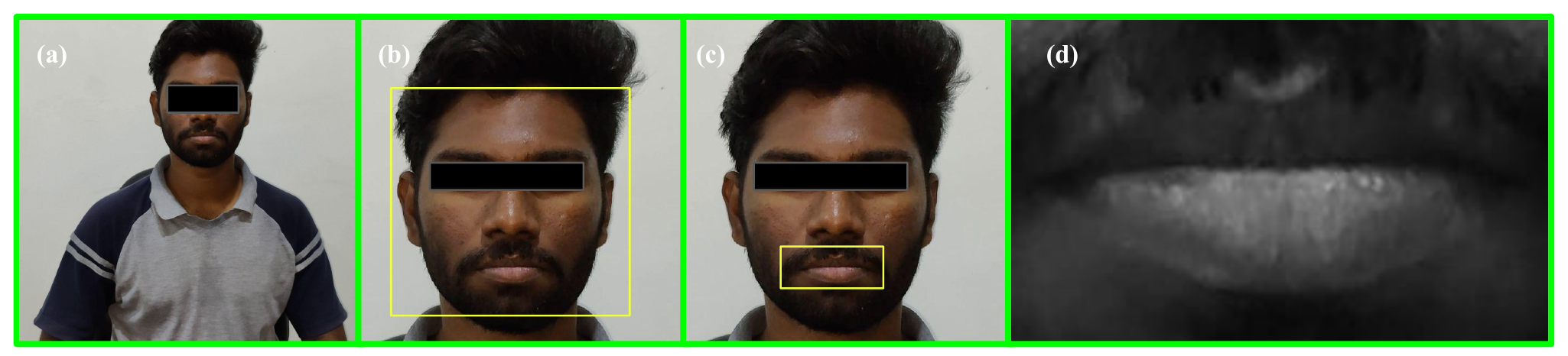}
    \caption{(a) Input Video Frame, (b) Face Localization, (c) Lip Region Cropping, (d) Feature Normalization and Gray Scale conversion. Credit: Nemani et al. \cite{nemani2022deep}}
    \label{fig:FN}
\end{figure}

\textbf{Li et al. \cite{9095331}} suggested a High-Fidelity Illumination Normalization for Face Recognition Based on Auto-Encoder. The proposed methodology involves the usage of conventional autoencoders to map the face under various illumination conditions and to perform feature normalization. The authors also claim that facial features are preserved, thus helping in effective facial recognition. Also, the methodology was evaluated on the CAS-PEAL database and the extended Yale B database and the results show the superior performance of the methodology on all angles and frames. Similar work involving the normalization of images under illumination conditions was proposed by \textbf{Ling et al. \cite{9344868}} The technique integrates GANs \cite{pan2019recent, wang2017generative, saxena2021generative, alqahtani2021applications} with feature maps of different sizes derived from pretrained feature networks using various convolutional layers and then uses these feature maps to compute loss. Under various lighting conditions, the suggested technique produces favorable illumination normalization results compared to earlier methods. Fig. \ref{fig:FN} depicts the phase of preprocessing and feature normalization. After the preprocessing step, this feature vector is then given as input to the speaker-independent VSR architectures, which are designed to recognize the spoken words regardless of the speaker's identity. In the following section, we discuss the different architectures used for the task of speaker independent VSR. 

\section{Proposed Methodologies for VSR}

\subsection{Summary of Seminal Works in VSR in the 1990s}
While the concept of VSR is not new, it gained significant attention in the 1990s due to advances in computer vision (CV) and ML. During this time, researchers developed various techniques and algorithms to analyze video recordings of speakers and recognize speech from visual input. These early studies laid the foundation for modern VSR systems and continue to be a source of inspiration for current research in the field. In the context of the study, we conducted a comprehensive review of \textbf{fifteen seminal works} in the field of VSR that share fundamental similarities with the contemporary concept of VSR. The review aims to analyze and compare the approaches, techniques, and algorithms employed in these works to gain insights into the evolution and progress of VSR research over time. The findings of the review provide valuable information and guidance for researchers and practitioners working in the area of VSR. In 1991, Gelder et al. \cite{gelder1991face} investigated the cognitive relationship between face recognition and lip-reading abilities in individuals with autism. The study suggested that individuals with \textbf{Autism Spectrum Disorder (ASD)} have difficulty recognizing faces but may be able to read lips at a level comparable to typically developing individuals. In 1993, works by Walden et al. \cite{walden1993benefit} and Silsbee et al. \cite{silsbee1993audio} illustrated that visual cues provide significant benefits in auditory-visual speech recognition by improving the accuracy of speech perception, especially in noisy environments. When speech is accompanied by visual cues, listeners are better able to distinguish between similar-sounding phonemes, resulting in increased speech intelligibility. 

\begin{figure}[ht]
    \centering
    \includegraphics[width = \columnwidth]{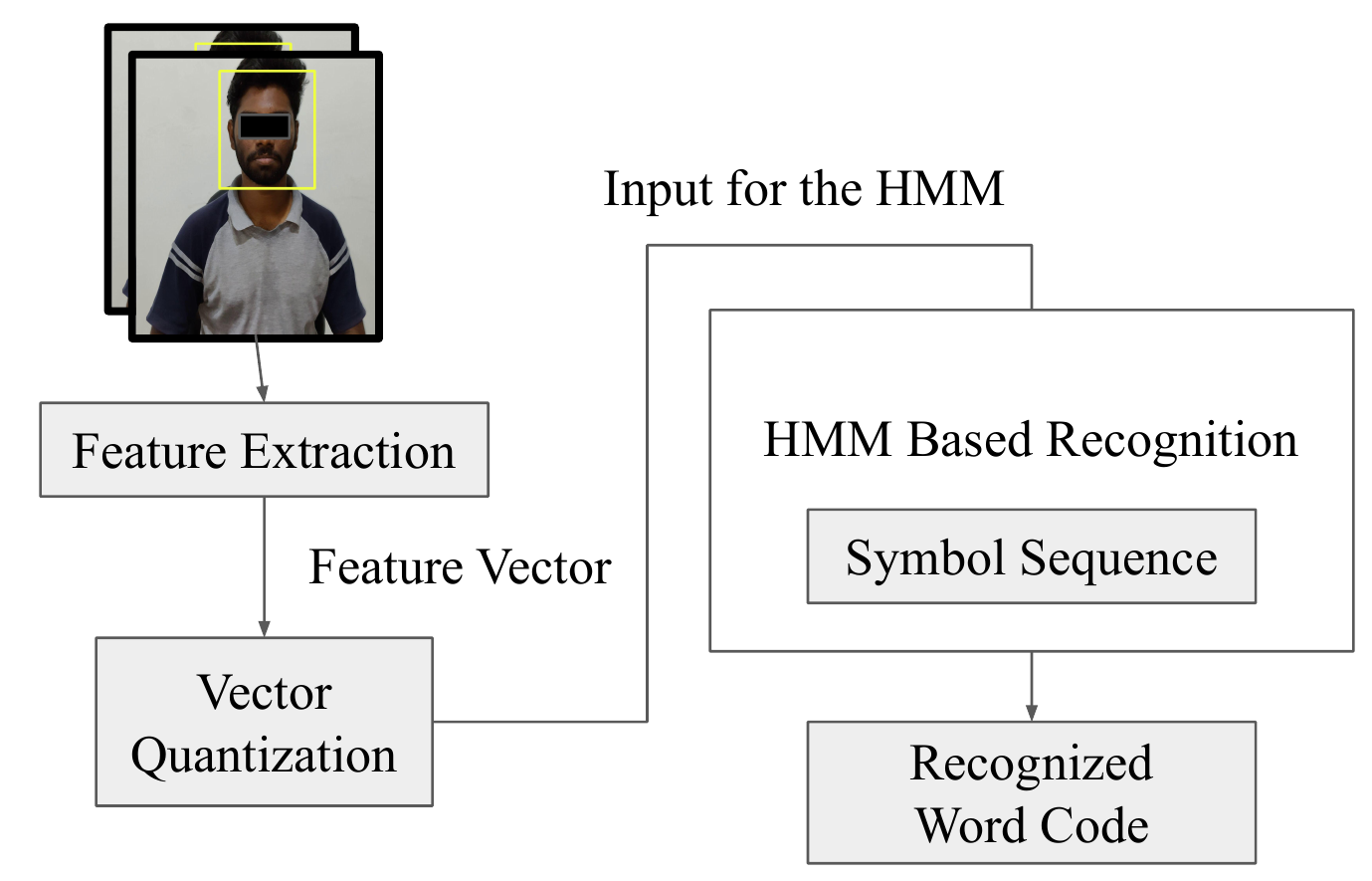}
    \caption{Underlying mechanism of HMMs}
    \label{fig:HMM}
\end{figure}

Visual cues can also help listeners to identify the speaker, as well as their emotional state, which can aid in understanding the meaning of the speech. This was further improved in 1994 when Javier et al. \cite{movellan1994visual} proposed a VSR methodology involving \textbf{stochastic networks}. The study described a speaker independent VSR system that used HMMs for limited vocabulary recognition. A \textbf{Hidden Markov Model (HMM)} \cite{bilmes2006hmms, jiang2010discriminative, young2008hmms} is a statistical model that is used to describe the probabilistic sequence of observable events, where the states generating the events are hidden or unknown. In an HMM, the observable events represent the outputs of a system, and the hidden states represent the underlying processes generating those outputs. The transitions between hidden states are governed by a set of probabilities called transition probabilities, which determine the probability of transitioning from one state to another. Additionally, each hidden state emits a set of observable events with a probability determined by a set of emission probabilities. The underlying mechanism of an HMM can be depicted in Fig. \ref{fig:HMM} The system recognizes the first four English digits and could be used in car-phone dialing applications. The images were modeled as mixtures of independent \textbf{Gaussian distributions}, and temporal dependencies were captured with left-to-right hidden Markov models. The results showed that the system achieved performance levels comparable to untrained humans in recognizing the first four English digits. 

In a series of studies conducted in the mid-1990s, researchers investigated the potential of using limited facial movement exposure to recognize monosyllabic words and improve speech recognition systems based on visual features. \textbf{Marassa et al. \cite{marassa1995visual}} in 1995 utilized a new method that restricted facial movements in video sequences to only the \textbf{lips-plus-mandible region} and found that speech readers could recognize monosyllabic words using video sequences that provided only this limited information. Bregler et al. \cite{bregler1995nonlinear} introduced a novel approach for learning smooth nonlinear manifolds and applied it to various lip reading tasks. Their technique was capable of determining the surface structure and finding the closest manifold point to a query point, thus enhancing the performance of acoustic speech recognizers in noisy environments. In a subsequent study in 1996, \textbf{Luettin et al. \cite{luettin1996statistical}} presented a speechreading system that relied solely on visual features extracted from \textbf{grey-level image sequences} of a speaker's lips. The system utilized \textbf{Active Shape Models} \cite{cootes1998active, cootes2001active} to track the lip contours and extract visual speech information from their shape, which was then modeled using continuous-density HMMs. The authors found that speech-relevant information was present in low dimensional space and was relatively robust to variations in both inter- and intra-speaker variability. \textbf{Luettin et al. \cite{luettin1996visual}} also introduced an approach to visual speech recognition using a similar Active Shape Model to represent the shape of the mouth, with recognition tests yielding an accuracy of 85.42\% for a speaker-independent recognition task of the first four digits using lip shape information only. Overall, these studies suggest that utilizing limited facial movement exposure and visual features can potentially enhance speech recognition systems.

In 1997, researchers developed innovative approaches to lipreading systems for speech recognition tasks using only visual information from human lips without any acoustic data. \textbf{Chiou et al. \cite{chiou1997lipreading}} presented a system that utilized snake algorithms to extract geometric visual features, \textbf{Karhunen-Lo`eve} transform to extract principal components in the color eigenspace, and HMMs to recognize visual feature sequences, achieving an impressive 94\% accuracy rate for ten isolated words. \textbf{Yu et al. \cite{yu1997lipreading}} proposed a method that treated the intensity of each pixel in an image sequence as a function of time, and lip movements were modeled by applying a \textbf{one-dimensional Fourier Transform} to this intensity-versus-time function. The approach was evaluated through experiments on two distinct databases of ten English digits and letters. In another study, \textbf{Luettin et al. \cite{luettin1997towards}} introduced a novel approach to speechreading that utilized visual feature extraction for speaker-independent continuous digit recognition. The lip tracker was used to extract information about the lip shape and the grey-level intensity around the mouth, which were then used to train visual word models using \textbf{Continuous-Density HMMs}. The experimental results showed that the method generalized well to new speakers, and the recognition rate varied considerably across digits due to the high visual confusability of certain words.

\begin{table*}[htbp]
	\centering
	\caption{Tabulated Overview of VSR Research from the 1990s}
	\label{Tab:1990}
	\resizebox{\linewidth}{!}{
	\begin{tabular}{|c|c|c|p{59mm}|c|}
	\hline
	\textbf{Existing Works}& 
	\centering \textbf{Year}& 
	\centering \textbf{Methodology}& 
	\centering \textbf{Dataset}& 
	\centering \textbf{Nature of the corpus} 
\tabularnewline
\hline

Gelder et al. \cite{gelder1991face} & 
1991 & 
- & 
\centering - & 
-
\tabularnewline
\hline

Walden et al. \cite{walden1993benefit} &
1993 & 
Visual Cue Derivation &
\centering Customized Dataset of 20 middle-aged and 20 elderly Subjects & 
Vowels and Sentences \tabularnewline 
\hline

Silsbee et al. \cite{silsbee1993audio} &
1993 & 
Visual Cue Derivation &
\centering - & 
Vowels \tabularnewline 
\hline

Javier et al. \cite{movellan1994visual} & 
1994 & 
Stochastic Networks & 
\centering Customized Dataset of 9 males, 3 females pronouncing 4 digits & 
Digits \tabularnewline
\hline

Marassa et al. \cite{marassa1995visual} & 
1995 & 
lips-plus-mandible region restriction &
\centering Customized Dataset of 26 normal hearing college students and 4
adults with bilateral sensorineural hearing loss &
Monosyllabic Words \tabularnewline
\hline

Bregler et al. \cite{bregler1995nonlinear} & 
1995 & 
HMM + Learned Lip Manifold &
\centering Customized Dataset of 4500 images of 6 speakers uttering random words &
Words \tabularnewline
\hline

Luettin et al. \cite{luettin1996statistical} &
1996 &
Active Shape Models + HMMs &
\centering Tulips1 &
Digits \tabularnewline
\hline

Luettin et al. \cite{luettin1996visual} &
1996 &
Improved Active Shape Models &
\centering Tulips1 &
Digits \tabularnewline
\hline

Chiou et al. \cite{chiou1997lipreading} &
1997 &
Karhunen-Lo`eve transform + HMM &
\centering Custom Dataset &
Continuous Digits \tabularnewline
\hline

Yu et al. \cite{yu1997lipreading} &
1997 &
Fourier Transforms &
\centering - &
- \tabularnewline
\hline

Luettin et al. \cite{luettin1997towards} &
1997 &
HMM &
\centering M2VTS Database &
Continuous Digits \tabularnewline
\hline

G. Rabi et al. \cite{614788} &
1997 &
RNNs &
\centering Custom &
Words \tabularnewline
\hline

Nanaku et al. \cite{nankaku1999intensity} &
1999 &
Continuous Density HMMs &
\centering Tulips1 &
Digits \tabularnewline
\hline

Javier et al. \cite{movellan1999diffusion} &
1999 &
Stochastic version of RNNs &
\centering Tulips1 &
Digits \tabularnewline
\hline

Baldwin et al. \cite{baldwin1999automatic} &
1999 &
Fuzzy Set Theory &
\centering Custom and Tulips1 &
Digits \tabularnewline
\hline

\end{tabular}}
\end{table*}

\textbf{Recurrent Neural Networks (RNNs)} \cite{mandic2001recurrent, lipton2015critical} are a type of neural network that is designed to handle sequential data. Unlike traditional neural networks, which only operate on a fixed-sized input, RNNs can process inputs of variable length, making them well-suited for tasks such as language modeling, speech recognition, and time series analysis. The key feature of an RNN is its ability to maintain a "hidden state" that captures information about the sequence processed so far. This hidden state is updated at each time step, allowing the network to capture long-term dependencies in the data. However, they suffer from some limitations, such as difficulties in handling long-term dependencies and a tendency to forget earlier inputs as the sequence gets longer. In this context, \textbf{G. Rabi et al. \cite{614788}} proposed a method for speech recognition based on RNNs, which involves first extracting time-varying visual speech patterns from a sequence of images, followed by classifying the spatiotemporal pattern as one of the previously trained words using recurrent neural networks. The recurrent network is trained using only feed-forward complexity by specifying a certain behavior when it is given exemplar sequences. The training sequences of a given word are segmented using adaptive segmentation.

In 1999, \textbf{Nanaku et al. \cite{nankaku1999intensity}} presented a novel approach to parameter estimation for continuous density HMMs in the context of visual speech recognition. The authors propose an average-intensity and location-normalized training method, which integrates the normalization process into model training. The proposed method is based on a maximum likelihood formulation, providing a theoretically well-defined algorithm, and the likelihood for the training data is guaranteed to increase at each iteration of the normalized training. \textbf{Javier et al. \cite{movellan1999diffusion}} introduced a novel approach to image sequence recognition that differs from HMMs. The proposed method is based on a stochastic version of recurrent neural networks, which the authors call diffusion networks. Unlike HMMs, diffusion networks use continuous state dynamics and generate continuous paths, which can be beneficial in computer vision tasks that require continuity. The paper presented a review of the necessary results for the implementation of diffusion networks and applies them to a VSR task. The experimental results demonstrate that diffusion networks outperform the results obtained with the best HMMs. \textbf{Baldwin et al. \cite{baldwin1999automatic}} presented a novel approach to automatic computer lip-reading using fuzzy set theory and mass assignment theory for feature extraction from video sequences. The proposed method utilizes simple rules based on fuzzy sets to generate a knowledge base for phonemes or sounds using probabilistic grid models. The system was trained and tested on phonemes from a medium-sized vocabulary of words and achieved a reasonable accuracy for classification. The method was further evaluated on the Tulips1 database and demonstrated efficient and general learning techniques that can be applied to different speakers. 

The analysis and comparison of the fifteen seminal works in the field of VSR presented in Table \ref{Tab:1990} highlight some significant findings and results. One of the key observations is that the majority of VSR solutions utilized HMMs for speech recognition. In particular, the works of Bregler et al. \cite{bregler1995nonlinear} and Luettin et al. \cite{luettin1996visual} demonstrated improved performance in speech recognition through the use of HMMs with learned lip manifolds and active shape models, respectively. However, towards the late 90s, modern deep learning methodologies like RNNs were also utilized in VSR research, as seen in the works of Javier et al. \cite{movellan1994visual} and G. Rabi et al. \cite{614788}. Another noteworthy observation is the utilization of standardized datasets such as Tulips1 and M2VTS in VSR research towards the late 90s. This allowed for more accurate and fair comparisons of different VSR techniques and served as a benchmark for future research. Despite the progress made in VSR research during the 90s, the performance metrics used in the majority of the solutions were not clear enough. Therefore, more work was required to establish better performance metrics and benchmarking procedures for VSR research. It is also important to note that not all works in the table explicitly mentioned the concept of speaker independence. This may be because some of the works focused on recognizing speech from a limited set of speakers, making speaker independence less relevant. Nonetheless, several works in the table addressed the issue of \textbf{speaker independence}. For instance, Bregler et al. \cite{bregler1995nonlinear} used a customized dataset of six speakers to train their system but tested it on a new speaker outside the training set. Similarly, Chiou et al. \cite{chiou1997lipreading} achieved speaker independence in the recognition of continuous digits using the Karhunen-Lo`eve transform and HMMs. The concept of speaker independence was recognized as an important challenge that needed to be tackled to develop a robust and practical VSR system. Overall, the works reviewed in the table provide a foundation for subsequent research in VSR, laying the groundwork for the development of more accurate and efficient VSR techniques in the future.

\subsection{Works between 2000 and 2010}

During the 1990s, traditional methodologies were the norm for speech recognition tasks, with little emphasis on developing models that could operate speaker-independently. However, with the advent of the new millennium, there was a shift towards evaluating traditional architectures on benchmark datasets, and the need for developing more speaker-independent models became more apparent. Towards the end of the 2000s, deep learning methodologies, such as \textbf{Artificial Neural Networks (ANNs)}, started gaining more attention in the VSR community. This approach allowed for more flexibility and better performance on VSR tasks. Despite the improved performance, the aspect of speaker independence was still less explored compared to the works after 2010. In this section, we review 20 such papers that describe the transition of VSR works from traditional methods to modern-day methodologies, with a specific focus on the aspect of speaker independence. 

In 2000, \textbf{Vanegas et al. \cite{vanegas2000lip}} presented a novel approach for enhancing the performance of a visual-information-based speech recognition system by normalizing the lip position. The proposed method aimed to address the issues that arise due to the variability in the lip location, which can adversely affect recognition accuracy. The lip location normalization algorithm is integrated into the model training process, and a search algorithm is employed to locate the lips' position accurately. Experiments were conducted on two speaker-independent databases, Tulips1 and M2VTS, for isolated word recognition. \textbf{The results demonstrate a recognition rate of 74.5\% and 94.8\% for the M2VTS and Tulips1 datasets, respectively, and an error reduction rate of 35.7\% and 76.3\% on the M2VTS and Tulips1 datasets, respectively}. \textbf{Nankaku et al. \cite{nankaku2000normalized}} proposed a novel approach for estimating the parameters of continuous density HMMs in VSR. One of the significant challenges in image-based visual speech recognition is the normalization of lip location and lighting conditions, which are crucial for the accurate estimation of HMM parameters. To address this issue, the paper presents a normalized training method that integrates the normalization process into the model training. This approach has been extended to include contrast normalization in addition to average intensity and location normalization. The proposed method utilizes a maximum likelihood formulation, which provides a theoretically well-defined algorithm. This formulation ensures that the likelihood of the training data increases at each iteration of the normalized training. Experiments have been conducted on the M2VTS database achieving a \textbf{recognition rate of 74.1\% and an error reduction rate of 24.7\%}, and the results demonstrate that the proposed approach can significantly improve recognition performance. The normalized training method provides a practical solution to address the challenges in VSR, and it has the potential to enhance the accuracy and robustness of the system. 

In 2002, numerous researchers utilized \textbf{Support Vector Machines (SVMs)} \cite{christmann2008support} in the field of VSR. SVMs are powerful and popular machine learning algorithms used for classification and regression tasks \cite{mavroforakis2006geometric}. SVMs are a supervised learning method that can be used for both linear and non-linear data classification. The basic idea of SVM is to find a hyperplane in a high-dimensional space that separates different classes of data points in such a way that the margin between the hyperplane and the closest data points is maximized. This margin is known as the maximum margin hyperplane (MMH). The MMH is found by solving a convex optimization problem that involves maximizing the distance between the hyperplane and the closest data points, subject to some constraints. One of the key advantages of SVMs is their ability to handle high-dimensional data sets and their ability to generalize well to new data points. SVMs are also less prone to overfitting than other classification algorithms, as they optimize the margin between the hyperplane and the closest data points rather than minimizing the classification error. In this context, \textbf{Gordan et al. \cite{gordan2002support}} investigated the feasibility of employing SVMs in VSR by modeling each word as a temporal sequence of visemes that correspond to different phones. The proposed method trained an SVM for each viseme to recognize and convert its output into a posterior probability through sigmoidal mapping. To capture the temporal aspect of speech, the SVMs were incorporated as nodes in a Viterbi lattice. The proposed approach is evaluated on a small VSR task, specifically the recognition of the first four English digits. The experimental results on the Tulips1 dataset depict a \textbf{word recognition rate of 90.6\%} on the addition of delta features,  demonstrating that the proposed method performs comparably to previously reported state-of-the-art results in terms of word recognition rate. The authors also compare the results with several works involving the usage of AAMs and HMMs and demonstrate the superior performance of the methodology in terms of word recognition rate and accuracy. The authors of the aforementioned paper have also proposed diverse modifications of SVMs for VSR, as outlined in \cite{gordan2002temporal} and \cite{1028281}.

In 2003, \textbf{Foo et al. \cite{1202350}} presented a novel boosted classifier for VSR that employed multiple HMMs. The composite HMMs are specifically trained to emphasize certain groups of training samples through the adaptive boosting technique. Experimental evaluations are conducted on a \textbf{customized dataset defined in the MPEG-4 multimedia standards} to identify the fundamental visual speech elements in English using the proposed boosted classifier. By comparing the outcomes of the proposed classifier with those of the traditional single HMM classifier, the system achieves an \textbf{accuracy of 37\%}, and it can be concluded that the proposed system is considerably superior in terms of accuracy and robustness, which demonstrates its potential for improving VSR performance. \textbf{Yao et al. \cite{yao2003visual}} presented a novel approach to selecting and extracting visual features for lipreading, which combines both low-level and high-level features that complement each other. The resulting feature set consists of 41 dimensions and is used for recognition. The approach is evaluated on a bimodal database called \textbf{AVCC}, which includes sentences that cover all Chinese pronunciations. The results show that the proposed method achieves \textbf{an accuracy of 87.8\%} for automatic speech recognition by lipreading assistance, which is an improvement from 84.1\%. \textbf{Moreover, it improves accuracy from 31.7\% to 51.2\% for \textbf{speaker-dependent recognition} and from 27.6\% to 55.3\% for speaker-independent recognition in noisy conditions}. The paper also demonstrates that visual speech information can effectively compensate for the loss of acoustic information caused by noise, with an \textbf{improvement rate ranging from 10\% to 30\%} depending on the amount of noise in speech signals. The improvement rate achieved by the proposed system is higher than that of the ASR system of IBM, and it performs better in noisy environments. 

In 2004, \textbf{Anwar et al. \cite{anwar2004learning}} proposed a method to learn fuzzy rules for VSR. The system automatically extracts features from video sequences and constructs a rule base using two-dimensional fuzzy sets on feature and time parameters. The method was applied to the \textbf{Tulips1 database}, and the results were better than those obtained with neural networks and Hidden Markov Models, implying the concept of \textbf{speaker independence}. \textbf{A medium-sized vocabulary of around 300 words}, representative of phonemes in the English language, was used for training and testing, achieving reasonable accuracy for phoneme classification. The \textbf{accuracy achieved was 21-33\%}, comparable to expert human lip-readers whose accuracy on nonsense words is about 30\%. \textbf{Foo et al. \cite{foo2004recognition}} introduced a new method for VSR that combines adaptive boosting and hidden Markov models to create an \textbf{AdaBoost-HMM classifier}. The approach trains composite HMMs to cover different groups of training samples and uses a novel probability synthesis rule to combine the decisions of component classifiers, resulting in a more complex decision boundary than the traditional single HMM classifier. The method is evaluated on a \textbf{customized dataset sampled at 50 frames per second with the typical viseme length of 0.3-1s} for recognition of basic visual speech elements and outperforms the traditional HMM classifier in accuracy, particularly for visemes extracted from contexts. \textbf{The proposed methodology achieves a training error of 10\% and a classification error of 16\%}. 

In 2005, \textbf{Saenko et al. \cite{1544886}} proposed an architecture for VSR that utilizes \textbf{discriminative detection of visual speech and articulate features}. The approach uses discriminative classifiers to identify the subclass of lip appearance corresponding to the presence of speech and decompose it into physical components of articulate production. These components evolve semi-independently, which conventional viseme-based approaches fail to capture. The authors propose a dynamic Bayesian network with a multi-stream structure to model co-articulation effects. The system is evaluated on a command utterance task that happens to be \textbf{a subset of the  AVTIMIT dataset} and shows promising results in lip detection, speech/non-speech classification with \textbf{an accuracy of 67\%}, and recognition performance against several baseline systems. \textbf{Sagheer et al. \cite{1577194}} proposed a new approach for VSR using a hypercolumn model (HCM) for feature extraction and HMM for modeling the extracted features with Gaussian distributions. The system is evaluated on an Arabic database set, and the accuracy achieved for words is 74\% and for sentences 55\%,  making it the first time VSR is applied to the Arabic language. The performance of HCM is compared to the fast discrete cosine transform (FDCT) approach using the same dataset and experimental conditions. Results show that HCM achieves higher recognition accuracy than FDCT and can achieve shift-invariant recognition. In 2006, \textbf{Lee et al. \cite{lee2006training}} introduced a new training algorithm for HMMs used in VSR. The algorithm is based on a modified version of \textbf{Simulated Annealing (SA) called hybrid simulated annealing}, which combines SA with a local optimization technique to improve both convergence speed and solution quality. Unlike the expectation-maximization (EM) algorithm, a popular HMM training method that only achieves local optima in the parameter space, the proposed algorithm performs a global search and thus obtains solutions that lead to better recognition performance. The algorithm is evaluated through isolated word recognition experiments, and \textbf{it displayed an error rate of 33.4\%}, thus demonstrating its effectiveness in improving recognition accuracy.

\textbf{Yau et al. \cite{1663790}} presented a novel approach for speech recognition using visual information by representing the speaker's mouth movement through \textbf{Motion History Images (MHIs)}. The MHIs are generated by applying accumulative image differencing on the video frames to capture temporal information. The decomposed wavelet sub-images obtained through discrete stationary wavelet transform (SWT) are used to extract three moment-based features (geometric moments, Zernike moments, and Hu moments). The features are then classified using a multilayer perceptron (MLP) \cite{taud2018multilayer, bisong2019multilayer} type ANN with a backpropagation learning algorithm. The paper compares and evaluates the classification ability of the different moments. The initial results demonstrate \textbf{an error rate of less than 5\% or an average accuracy of 95\%} for English consonant classification on the \textbf{created customized dataset}. In 2007, \textbf{Leung et al. \cite{1699666}} introduced an automatic lipreading technique for both \textbf{speaker dependent and speaker independent speech recognition tasks}. The approach employs a spline representation to convert the visual features, which are extracted based on the video sequence frame rate, into the continuous domain. To improve the accuracy of the recognition system, the spline coefficients in the same word class are constrained to have similar expressions and are estimated using the \textbf{Expectation-Maximization (EM)} algorithm on the training data. Moreover, the paper proposes an adaptive multi-model approach to account for the variation in speaking style during the speaker-independent recognition task. The experimental results demonstrate the effectiveness of the proposed approach, achieving \textbf{an accuracy of 96\% for speaker dependent recognition} and \textbf{88\% for speaker independent recognition} on recognizing ten English digits. The results also indicate that the proposed approach outperforms other classifiers investigated in the study. 

\begin{table*}[ht]
	\centering
	\caption{Examining VSR works from 2000 to 2010: a tabulated overview}
	\label{Related}
	\resizebox{\linewidth}{!}{
	\begin{tabular}{|c|c|p{45mm}|c|c|p{50mm}|}
	\hline
	\centering \textbf{Existing Works}& 
	\centering \textbf{Year}& 
	\centering \textbf{Methodology}& 
	\centering \textbf{Dataset}& 
	\centering \textbf{Nature of Corpus}& 
	\centering \textbf{Performance Metrics} 
\tabularnewline
\hline

\centering Vanegas et al. \cite{vanegas2000lip} &
\centering 2000 &
\centering Lip Position Normalization + Model Training &
\centering Tulips1 &
\centering Digits &
\centering Recognition Rate = 94.8\% , Error Reduction Rate = 76.3\%
\tabularnewline
\hline

\centering Vanegas et al. \cite{vanegas2000lip} &
\centering 2000 &
\centering Lip Position Normalization + Model Training &
\centering M2VTS &
\centering Digits &
\centering Recognition Rate = 74.5\% , Error Reduction Rate = 35.7\%
\tabularnewline
\hline

\centering Nankaku et al. \cite{nankaku2000normalized} &
\centering 2000 &
\centering Continous HMMs &
\centering M2VTS &
\centering Digits &
\centering Recognition Rate = 74.1\% , Error Reduction Rate = 24.7\%
\tabularnewline
\hline

\centering Gordan et al. \cite{gordan2002support} &
\centering 2002 &
\centering SVMs &
\centering Tulips &
\centering Digits &
\centering Recognition Rate = 90.6\%
\tabularnewline
\hline

\centering Foo et al. \cite{1202350} &
\centering 2003 &
\centering Adaboost + HMM &
\centering Custom &
\centering Words &
\centering Accuracy = 37\%
\tabularnewline
\hline

\centering Yao et al. \cite{yao2003visual} &
\centering 2003 &
\centering Low-level and high-level feature combination &
\centering AVCC &
\centering Sentences &
\centering Accuracy = 55.3\% for speaker independent VSR
\tabularnewline
\hline

\centering  Anwar et al. \cite{anwar2004learning} &
\centering 2004 &
\centering Fuzzy Logic &
\centering Tulips1 &
\centering Digits &
\centering Accuracy = 33\% 
\tabularnewline
\hline

\centering Foo et al. \cite{foo2004recognition} &
\centering 2004 &
\centering AdaBoost-HMM Classifier &
\centering Custom &
\centering - & 
\centering Training Error = 10\%, Classification Error = 16\% \tabularnewline
\hline

\centering Saenko et al. \cite{1544886} &
\centering 2005 &
\centering Dynamic Bayesian Network &
\centering Subset of AVTIMIT &
\centering Sentences &
\centering Viseme Classification Accuracy = 63\% \tabularnewline
\hline

\centering Sagheer et al. \cite{1577194} &
\centering 2005 &
\centering hypercolumn model (HCM) + HMMs &
\centering Custom &
\centering Words, Sentences&
\centering Word Accuracy = 74\%, Sentence = 55\% \tabularnewline
\hline

\centering Lee et al. \cite{lee2006training} &
\centering 2006 &
\centering Simulated Annealing &
\centering Digit Database &
\centering Digits &
\centering Error Rate = 33.4\% \tabularnewline
\hline

\centering Yau et al. \cite{1663790} &
\centering 2006 &
\centering SWT + MLP &
\centering Custom &
\centering Consonants &
\centering Error Rate = 5\% \tabularnewline
\hline

\centering Leung et al. \cite{1699666} &
\centering 2007 &
\centering adaptive multi-model approach &
\centering Custom &
\centering Digits &
\centering speaker dependent accuracy = 96\% and  speaker independent = 88\% \tabularnewline
\hline

\centering Yu et al. \cite{yu2007new} &
\centering 2007 &
\centering KNN + HMM Classifier &
\centering Custom &
\centering Words &
\centering Avg Accuracy = 95\% \tabularnewline
\hline

\centering Jie et al. \cite{4712008} &
\centering 2008 &
\centering Modular Bidirectional PCA (MBDPCA) &
\centering CUAVE &
\centering Digits &
\centering Accuracy  = 95.71\% \tabularnewline
\hline

\centering Wang et al. \cite{4667054} &
\centering 2008 &
\centering Genetic Algorithm &
\centering Custom &
\centering Vowels &
\centering Accuracy = 91.47\% \tabularnewline
\hline

\centering Rajavel et al. \cite{rajavel2009static} &
\centering 2009 &
\centering DCT + Motion Segmentation-based Dynamic Feature Approach & 
\centering Custom &
\centering Consonants &
\centering Accuracy = 92.45\%, 92.15\% for both approaches \tabularnewline
\hline

\centering Pass et al. \cite{5652630} &
\centering 2010 &
\centering Contexual Modeling using HMMs &
\centering XM2VTS &
\centering Phrases &
\centering Improvement of over 17\% relative Word Error Rate \tabularnewline
\hline

\centering Lu et al. \cite{5713068} &
\centering 2010 &
\centering Convolutional VEF snake &
\centering Custom &
\centering Digits, Words &
\centering Average Recognition Accuracy = 92.3\% \tabularnewline
\hline

\end{tabular}}
\end{table*}

\textbf{Yu et al. \cite{yu2007new}} presented a novel approach for VSR by proposing a new manifold representation. The real-time input video data is compressed using \textbf{Principal Component Analysis (PCA)} to define manifolds based on the low-dimensional points calculated for each frame. Since the number of frames varies based on the complexity of the word, the manifolds need to be resampled into a fixed number of key points for visual speech classification. Two classification schemes, the k Nearest Neighbour (kNN) algorithm with two-stage PCA and HMM classifier, are evaluated on a \textbf{customized dataset} with the nature of the corpus being \textbf{words}. The proposed approach is shown to produce accurate classification results for a group of English words with an \textbf{average classification of 95\%}. In 2008, \textbf{Jie et al. \cite{4712008}} proposed a new approach to feature extraction called \textbf{Modular Bidirectional PCA (MBDPCA)}, which is a modification of Bidirectional PCA (BDPCA). The MBDPCA method divides the original mouth frames into smaller sub-frames and uses two approaches for building the covariance matrix. The first approach builds a global covariance matrix using all sub-frame sets collectively, while the second approach builds local covariance matrices using various sub-frame sets separately. Afterward, each sub-frame set is processed by BDPCA. Their study and experimentation on the \textbf{CUAVE Dataset} showed that MBDPCA outperformed PCA and BDPCA in accuracy, with the former achieving \textbf{an accuracy of 95.71\%}. However, the experimentation was limited to a small vocabulary dataset. In a different study, \textbf{Wang et al. \cite{4667054}} proposed a speech recognition technique that uses visual features and HMM. In their technique, the underlying state process of the HMM, which is considered a Markov mesh, has transition probabilities that depend on the states of surrounding blocks in both horizontal and vertical directions, treating feature vectors as statistically dependent. As a result, the dependence in two dimensions is simultaneously reflected. To train the HMM, a new \textbf{Genetic Algorithm (GA)} \cite{lambora2019genetic, mirjalili2019genetic} was proposed for global optimization. The study used ten disabled speakers as subjects and six Chinese vowels were used for testing. The improved HMM model resulted in an \textbf{average recognition rate increase of 2.51} and \textbf{an accuracy of 91.47\%} compared to the traditional HMM model.

In 2009, researchers at the \textbf{Delft University of Technology} recorded a custom dataset to build their system \cite{10.1007/978-3-642-15760-8_33}. The study utilized three different image processing methods, including \textbf{Active Appearance Models (AAMs)} \cite{cootes1998active, cootes2001active} for landmark recognition, lip geometry estimation to locate all lip pixels using a color filter, and optical flow for motion recognition in a visual representation. The HMM and Hidden Markov Model Toolkit (HTK toolkit) were employed for actual recognition. To detect the face region, the input visual speech video was given to the face localization module, and the mouth region was identified regarding the face region. \textbf{Rajavel et al. \cite{rajavel2009static}} investigated the effectiveness of static and dynamic visual speech features for improving visual speech recognition. To extract visual features, two approaches are considered: (1) an image transform-based static feature approach that employs Discrete Cosine Transform (DCT) on each video frame and reduces redundancy using PCA, resulting in 21 static visual features, and (2) a motion segmentation-based dynamic feature approach that uses motion history images (MHI) to segment facial movements from the video and applies DCT on triangle region coefficients to obtain dynamic visual features. Two types of experiments are conducted using concatenated features and dimension-reduced features obtained through PCA. A left-right continuous Hidden Markov Model (HMM) is used to classify nine MPEG-4 standard viseme consonants. The experimental results demonstrate that both concatenated and dimension-reduced features enhance visual speech recognition accuracy, achieving \textbf{a high accuracy rate of 92.45\% and 92.15\%,} respectively. 

In 2010, \textbf{Pass et al. \cite{5652630}}  presented a novel approach to visual speech recognition that enhances contextual modeling by combining Inter-Frame Dependent and Hidden Markov Models. By incorporating contextual information, this approach is able to capture nuances in visual speech that may be missed by relying solely on a Hidden Markov Model. The proposed method is evaluated on a large speaker independent isolated digit recognition task and is compared against two commonly used feature-based techniques for incorporating speech dynamics. The results from baseline feature-based systems and the combined modeling approach are presented and analyzed. It is demonstrated that while both techniques perform similarly when used alone, a significant improvement in performance can be achieved by combining them. Specifically, the combined modeling approach yields an \textbf{improvement of over 17\% relative Word Error Rate} compared to the best baseline system. These findings highlight the potential of combining different modeling techniques to improve VSR. \textbf{Lu et al. \cite{5713068}} proposed a new approach for automatic visual speech recognition using \textbf{Convolutional VEF snake and canonical correlations}. The approach utilizes a head-mounted camera to record the utterance image sequences of isolated Chinese words and applies the Convolutional VEF snake model to detect and track lip boundaries in a rapid and accurate manner. Geometric and motion features are extracted from lip contour sequences and combined to form a joint feature descriptor. Canonical correlation is employed to measure the similarity of two utterance feature matrices, and a linear discriminant function is introduced to improve the recognition accuracy further. With an average recognition \textbf{accuracy of 92.3\% for digits and words combined}, the experimental results show that the proposed approach is promising, and the joint feature descriptor is more robust than individual ones.

\subsection{Works between 2011 and 2023}

In 2011, \textbf{Damien et al. \cite{damien2011visual}} proposed a novel approach to visual speech recognition which involves introducing a consonant-vowel detector and utilizing two classifiers: an HMM-based classifier for the recognition of the "consonant part" of the phoneme and a classifier for the "vowel part". This approach offers several benefits, such as reducing the number of hidden states and the number of HMMs. The proposed method was tested on a limited set of words in the Modern Classic Arabic language, and \textbf{a recognition rate of 81.7\%} was achieved. Additionally, the proposed model is speaker-independent and uses visemes as the basic units, making it applicable to any set of words with varying content or size. \textbf{Shaik et al. \cite{shaikh2011visual}} presented a novel lip-reading technique that can identify visemes using visual data only, without relying on the corresponding acoustic signals. The approach is based on analyzing the vertical components of the \textbf{optical flow (OF)}, which are classified using \textbf{support vector machines (SVM)}. To achieve automatic temporal segmentation of the utterances, the pair-wise pixel comparison method is used to evaluate the differences in intensity of corresponding pixels in two successive frames. The OF is decomposed into non-overlapping fixed scale blocks, and the statistical features of each block are computed for successive video frames of an utterance. The experiments were conducted on a database of 14 visemes taken from seven subjects, and the accuracy was tested using five and ten-fold cross-validation for binary and multiclass SVMs, respectively, to determine the impact of subject variations. With \textbf{an accuracy of 85\%}, the results indicate that the proposed method is more robust to inter-subject variations compared to other systems in the literature, with high sensitivity and specificity for 12 out of 14 visemes.

In 2012, \textbf{P. Sujatha et al. \cite{6179154}} tested eight different feature extraction techniques for speech recognition, and the 16-point Discrete Cosine Transform (DCT) method achieved the highest accuracy rate of 93.5 percent. The feature vectors obtained from the 16-point DCT method were given as input to the HMM, which was trained and tested on a dataset consisting of 10 speakers uttering 35 different words, and a \textbf{test accuracy of 97.5\% is obtained}. In 2013, Salah W et al. \cite{DBLP:journals/corr/abs-1301-4558} proposed an \textbf{Automatic Lip-reading Feature Extraction (ALiFE)} prototype for recognizing the vowels uttered by multiple speakers. The ALiFE system has three phases. In the first phase, the lip contour is identified and tracked, and a set of Points of Interest (POIs) is identified. In the second phase, the lip features are extracted, and the extracted visual data is categorized to identify the vowel sound. One of the challenges in visual-only speech recognition is controlling aspects such as lighting, identity, motion, emotion, and expression, but some aspects, such as video resolution, can be controlled. The experiments were conducted on a \textbf{customized dataset whose nature of corpus is vowels}, and a \textbf{training accuracy of 73.33\%} and \textbf{a testing accuracy of 72.73\%} was obtained. In 2014, A study by \textbf{Helen et al. \cite{7025274}} addressed the impact of resolution on speech recognition using a novel dataset called Rosetta Raven. The study employed the \textbf{Active Appearance Models (AAMs)} method for feature extraction, and the results suggested that resolution is not essential for automatic lip-reading. In their study, Helen et al. analyzed the impact of video resolution on automatic lip-reading using the Rosetta Raven dataset. The results of the study suggest that resolution may not be a crucial factor in speech recognition and that feature extraction methods such as AAMs can be employed to extract relevant features from low-resolution video data. Overall, these studies provide valuable insights into the challenges of visual-only speech recognition and propose effective solutions to address them.

\begin{figure}[htbp]
    \centering
    \includegraphics[width = \linewidth]{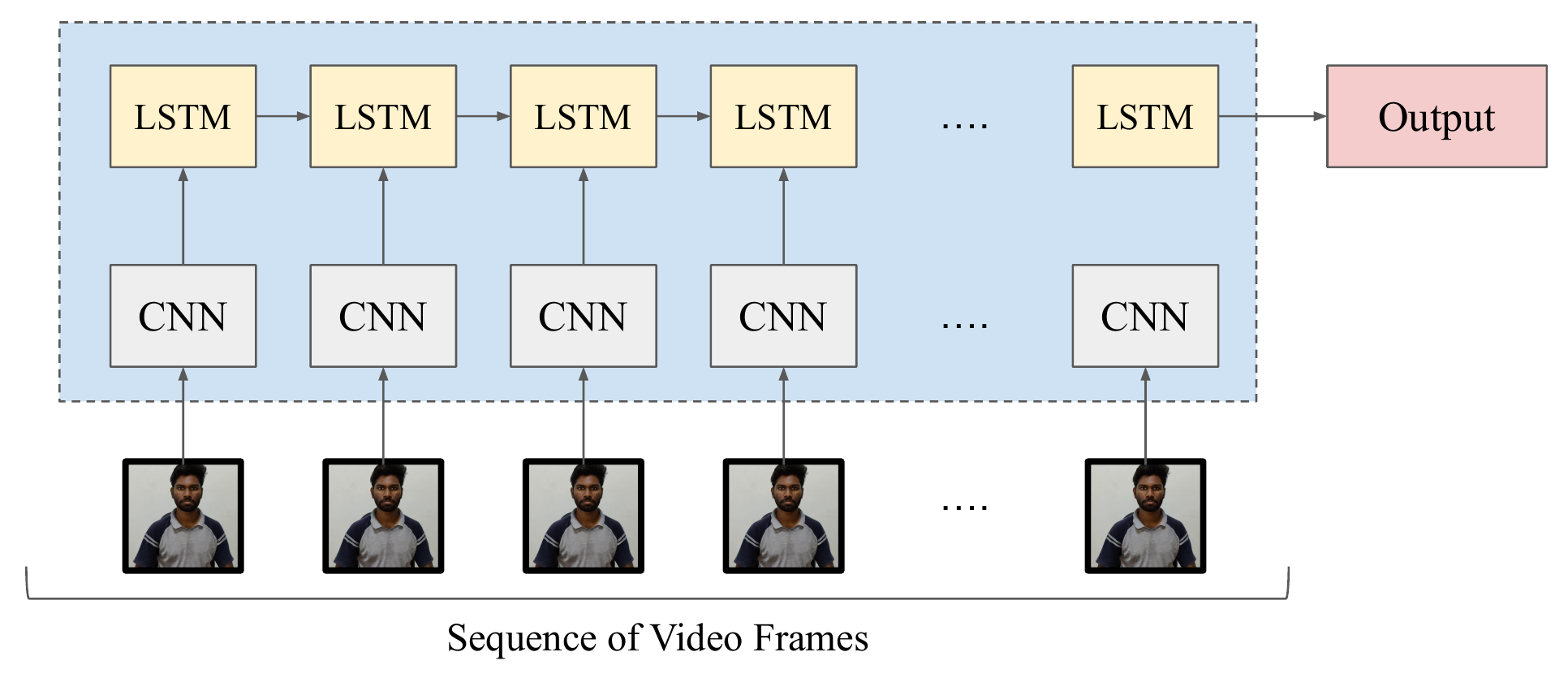}
    \caption{CNN and LSTMs combination for VSR}
    \label{fig:CNN+LSTMS}
\end{figure}

In 2016, \textbf{Amit et al. \cite{Amit2016LipRU}} proposed several techniques for predicting words and phrases from videos without any audio. The first approach involved combining a specific number of images from a video sequence to create a single image, which was then used as input to a CNN architecture, VGG Net. The VGG Net is a CNN architecture that was introduced in 2014 by the Visual Geometry Group (VGG) at the University of Oxford \cite{7486599, sengupta2019going, tammina2019transfer, koonce2021vgg}. It is a widely used CNN architecture for image classification tasks due to its simplicity and high accuracy. The VGG Net consists of 16 convolutional layers followed by 3 fully connected layers. Each convolutional layer has a fixed 3x3 filter size with a stride of 1 and padding of 1, and the max-pooling layer with a 2x2 filter size with a stride of 2 is applied after every two convolutional layers. The VGG Net architecture is characterized by its depth and uniformity, meaning that all the convolutional layers have the same filter size, and the number of filters is doubled after every pooling layer. The VGG Net was pre-trained on faces and served as a set of weights. The second method was similar to the first, but it expanded and normalized the number of images per sequence using nearest-neighbor interpolation. In the third approach, a single image was fed into the VGGNet to extract features, which were then passed into the LSTM layers \cite{6707742}. The proposed methodology achieved \textbf{a training accuracy of 66.15\%} and \textbf{a testing accuracy of 44.5\%} on the \textbf{MIRACL-VC1 phrases} corpus, thus making it one of the pioneering works experimented on the same. The combination of CNNs and Long Short-Term Memory (LSTM) networks has proven to be highly effective in this field, as depicted in Figure \ref{fig:CNN+LSTMS}.  \textbf{Chung et al. \cite{DBLP:journals/corr/ChungSVZ16}} proposed a primary methodology that uses Watch, Listen, Attend, and Spell (WLAS) architecture to translate recordings of mouth action into words. They developed a mechanism that handles audio and video input separately and together. Their work emphasized a curriculum-driven approach to increase training and minimize the issue of excessive variance. They trained the Weighted-Average Least Squares (WALS) model on the LRS dataset, which contains more than 10,000 sentences. Additionally, the WLAS architecture has been employed to translate mouth action recordings into words, which serves as a promising approach for speech recognition.

\begin{figure}[htbp]
    \centering
    \includegraphics[width = \columnwidth]{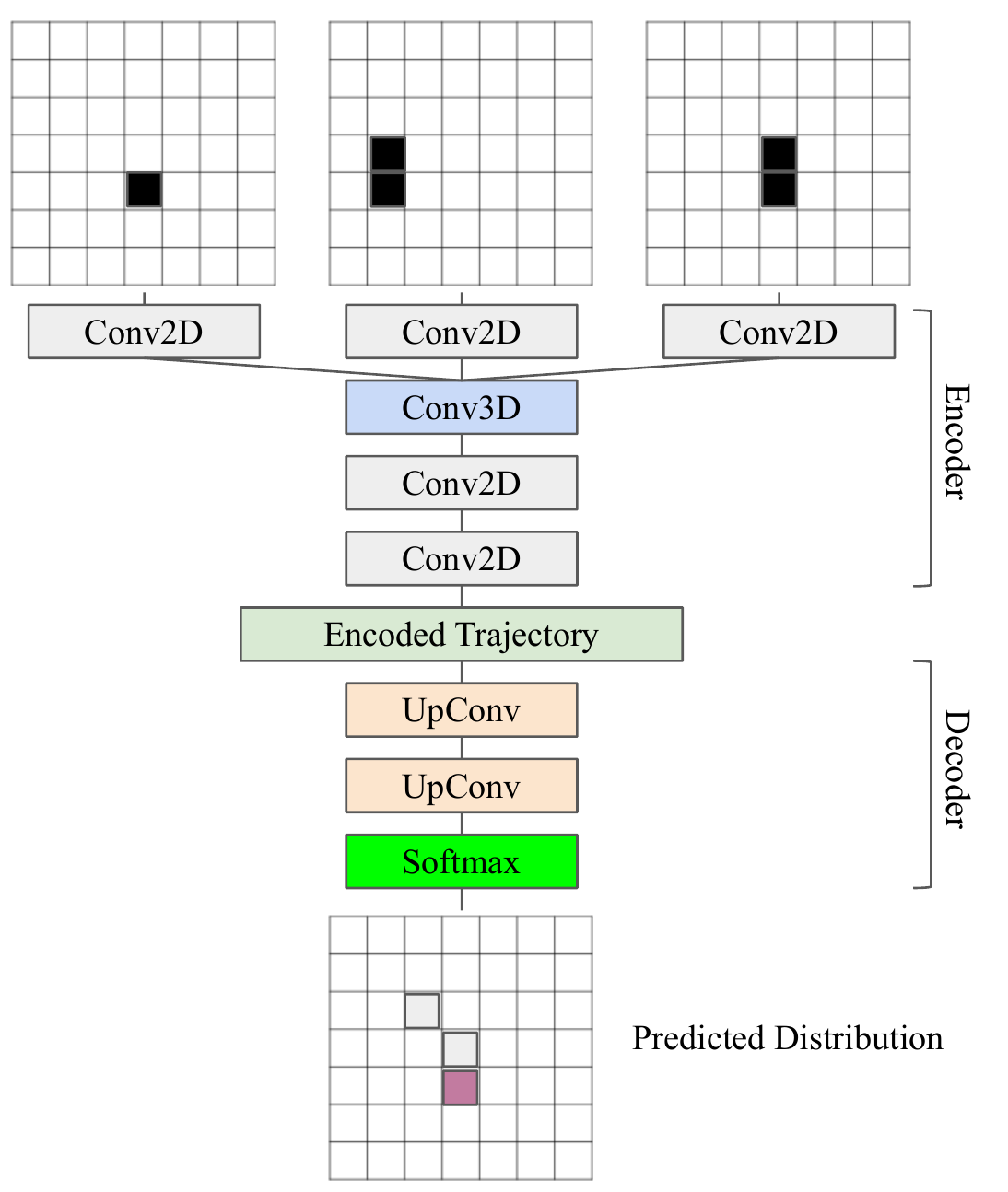}
    \caption{Architectural Mechanism of STCNNs}
    \label{fig:STCNN}
\end{figure}

\textbf{Assael et al. \cite{DBLP:journals/corr/AssaelSWF16}} proposed an end-to-end sentence-level VSR that converts sequences of picture frames of a speaker's mouth to complete sentences using \textbf{spatiotemporal convolutional neural networks (STCNNs)} \cite{he2019stcnn, zhao2019four, xu2019spatiotemporal} tested on the GRID corpus dataset. Spatiotemporal Convolutional Neural Networks (STCNNs) are deep learning models used for analyzing and classifying spatiotemporal data, such as video data. These models are an extension of the popular CNNs, which have proven to be highly effective in image recognition tasks. The key difference between STCNNs and traditional CNNs is that STCNNs can capture both spatial and temporal information from the data. This is achieved by using 3D convolutional layers, which convolve the input data along three dimensions (width, height, and time). The 3D convolutional layers are typically followed by pooling layers, which reduce the spatial resolution of the data, and recurrent layers, such as LSTMs or \textbf{Gated Recurrent Units (GRUs)} \cite{chung2014empirical}, which capture temporal dependencies between the frames in the video sequence. STCNNs have several advantages over traditional CNNs for video analysis tasks. Firstly, they can capture both spatial and temporal features simultaneously, allowing for more accurate and robust classification. Secondly, they can handle variable-length video sequences by using recurrent layers, which can process inputs of varying lengths. Finally, STCNNs can be trained end-to-end, meaning that the entire network can be optimized using backpropagation, which simplifies the training process. The core idea of an STCNN can be depicted in Fig. \ref{fig:STCNN}. A methodology is built to eliminate the requirement of dividing the visuals into words before predicting the phrase. Experimentations were conducted on the \textbf{LRS-Sentences corpus}, and \textbf{a testing accuracy of 95.2\%} was achieved. This methodology was compared with hard-of-hearing people’s ability to perform the task of VSR and significantly improved accuracy. 

Numerous research studies have utilized \textbf{Deep Complementary Bottleneck  Features} \cite{petridis2016deep, sainath2012auto, yu2011improved, zhang2015deep} from audio for speech recognition. However, the application of these features to video data is currently limited. To address this issue, \textbf{Stavros et al. \cite{7472088}} proposed a study utilizing deep encoders to build DBNFs. These deep encoders were trained with a bottleneck layer to reduce the image's dimensionality. Additionally, during training, DCT features were added to the bottleneck layer to make the bottleneck features complementary to the DCT features. The proposed method was evaluated on two well-known datasets, OuluVS and AVLetters, using an LSTM model, and the \textbf{accuracy achieved was 81.8\% for the letters corpus and 58.1\% for the sentence corpus}. In 2017, \textbf{Philip et al. \cite{8356433}} presented a novel method for liveness evaluation based on VSR. Their approach involved a combined system consisting of DNN and HMM models for speech recognition. The \textbf{XM2VTS dataset} was utilized for training and testing the model and the dataset achieved \textbf{a testing accuracy of 86.3\%}. To estimate liveness, the system used model VSR and calculated the Levenshtein Distance \cite{yujian2007normalized} between a randomly generated challenge phrase and the hypothesis utterances from the visual speech recognizer. \textbf{Eric et al. \cite{7952701}} proposed a system that combines a camera with an ultrasound imaging system to monitor the subject's lips and tongue movement. In this work, CNNs were employed to extract visual features from the original ultrasound and video images. The authors suggested a multimodal architecture that simultaneously processed two visual perceptions and developed various methods for using CNNs as a feature extractor combined with an HMM-GMM decoder. \textbf{The proposed methodology was evaluated on a customized dataset, and a testing accuracy of 80.4\% was achieved.} 

\textbf{AlexNet} is a deep CNN that won the ImageNet Large Scale Visual Recognition Challenge (ILSVRC) in 2012, marking a breakthrough in computer vision research. The network architecture consists of five convolutional layers, followed by three fully connected layers, and uses the Rectified Linear Unit (ReLU) activation function to avoid the vanishing gradient problem. AlexNet also incorporates techniques such as overlapping pooling, data augmentation, and dropout regularization to prevent overfitting \cite{9451239, krizhevsky2012imagenet}. The network was trained on over 1 million images from 1000 different classes and achieved a top-5 error rate of 15.3\%, significantly outperforming previous state-of-the-art methods. The success of AlexNet paved the way for further research and development in deep learning and its applications in computer vision. Similarly, \textbf{Inception V3} is a CNN architecture that utilizes inception modules \cite{xia2017inception, wang2019pulmonary, jignesh2020face}. These modules use parallel convolutional layers of different sizes to capture features at different scales. Inception V3 also incorporates batch normalization, factorized 7x7 convolutions, and auxiliary classifiers to improve training and regularization. The network has a deep architecture with 48 layers, including stem layers, inception modules, and fully connected layers. In a study by \textbf{P. Sindhura et al. \cite{9001505}} in 2018, AlexNet was thoroughly evaluated alongside Inception V3 for lip reading. Both models were trained using the MIRACL-VC1 dataset and were analyzed from both speaker-dependent and speaker-independent perspectives. The authors extracted each word's lip region individually and combined them into a single image, which was used to train the algorithm. \textbf{The experimentation was conducted on the MIRACL-VC1 words dataset, and an accuracy of 37.1\% was illustrated.}

\textbf{H. Gupta et al. \cite{8530509}} proposed a lip-reading model using CNN batch normalization for audio-less video data. The Haar Cascade algorithm is employed to extract the lip region from each individual frontal facial image in the video sequence and combine them into a single image. Haar Cascade is an object detection algorithm that is widely used in computer vision. It is based on the Haar wavelet technique and uses machine learning to detect objects in images or videos. The algorithm works by first extracting features from the input image using Haar-like features. These features are then used to train a classifier, such as the AdaBoost algorithm. The trained classifier is then used to detect objects in the input image. Haar Cascade has been successfully applied to face detection, pedestrian detection, and object recognition tasks. The algorithm is widely used in real-world applications and is known for its accuracy and speed. Furthermore, for extracting the visual features, a 12-layer CNN with two layers of batch normalization is used for training the model. The proposed model was trained using the MIRACL-VC1 dataset. The methodology depicted \textbf{a high training accuracy of 96.5\%, but, however, a low testing accuracy of 52.9\%}. In 2019, \textbf{M. A. Abrar et al. \cite{9042439}} developed a CNN model for predicting words from videos without any acoustic features. The videos are categorized into frames and are reduced to equivalent lengths, and the ROI is cropped from the videos. A pre-trained VGGNet architecture is trained on the  MIRACL-VC1 dataset. This methodology, too, achieved \textbf{a high training accuracy (94.86\%)} and \textbf{a low testing accuracy (60\%)}, implying the concept of overfitting. To operate the framework in real-time, an application has been developed to support the hearing impaired in their day-to-day activities. 

\textbf{D.Parekh et al. \cite{9033664}} proposed the use of convolutional auto encoders for extracting the feature vectors from the video frames, and these feature vectors are given as input to the LSTM model. Three separate standard datasets: BBC's LRW, MIRACL-VC1, and the GRID dataset, are used to test the proposed approach. \textbf{The model achieved an accuracy of 98\% and 63.22\% for speaker dependent and speaker independent on the MIRACL-VC1 dataset and an accuracy of 84.8\% on the GRID dataset. CNN is used as a baseline model for extracting the features.} Then the performance of the proposed model is compared with the baseline model. \textbf{Y. Lu et al. \cite{app9081599}} suggested three-step procedures for automatic lip-reading recognition. The first step is to extract the frames from the \textbf{created custom dataset} (Pronunciation of 0-9 digits by three men and three women in the English language). The feature vectors have been extracted from the frames using VGG Net. Then, to identify lip-reading, an attention-based LSTM-CNN fusion model was developed. \textbf{With a training accuracy of 88.2\% and a testing accuracy of 84.9\%}, the results obtained from the fusion attention-based model have been compared with the general RNN-CNN model, and it has shown that the proposed model has achieved more accuracy on the test data.

\begin{table*}[ht]
	\centering
	\caption{Comparison between existing VSR works}
	\label{Related}
	\resizebox{\linewidth}{!}{
	\begin{tabular}{|c|c|c|c|p{18mm}|p{15mm}|p{15mm}|}
	\hline

\centering \textbf{Existing Works} & 
\centering \textbf{Methodology} & 
\centering \textbf{Dataset} & 
\centering \textbf{Year} & 
\centering \textbf{Nature of Corpus} & 
\centering \textbf{Train Accuracy} &
\centering \textbf{Test Accuracy} \tabularnewline
\hline

\centering Damien et al. \cite{damien2011visual} &
\centering HMM Classifier &
\centering Custom &
\centering 2011 &
\centering Words &
\centering 81.67\% &
\centering - \tabularnewline
\hline

\centering Shaik et al. \cite{shaikh2011visual} &
\centering Optical Flow + SVM &
\centering Custom &
\centering 2011 &
\centering Vowels, Consonants &
\centering 85\% &
\centering -  \tabularnewline
\hline

\centering Sujatha et al. \cite{6179154} &
\centering DCT + HMM& 
\centering Custom & 
\centering 2012 & 
\centering Words & 
\centering -  & 
\centering 97.5\% \tabularnewline
\hline

\centering Salah W et al. \cite{DBLP:journals/corr/abs-1301-4558}&  
\centering POI + Energy & 
\centering Custom & 
\centering 2013 & 
\centering Vowels & 
\centering 73.33\%  & 
\centering 72.73\% 
\tabularnewline
\hline

\centering Helen L et al. \cite{7025274}& 
\centering HMM + HTK Toolkit
& \centering Rosetta Raven & 2014
& \centering 
& \centering - &
\tabularnewline
\hline

\centering Amit et al. \cite{Amit2016LipRU}&  \centering CNN + LSTM
& \centering MIRACL-VC1 & \centering 2016
& \centering Phrases
& \centering 66.15\% & \centering 44.50\%
\tabularnewline
\hline

\centering J. S. Chung et al. \cite{DBLP:journals/corr/ChungSVZ16}&  \centering CNN + LSTM
& \centering LRS & \centering 2016
& \centering Sentences
& \centering - & \centering -
\tabularnewline
\hline

\centering Assael et al. \cite{DBLP:journals/corr/AssaelSWF16}&  
\centering Spatio-Temporal CNN + GRU
& \centering LRS & \centering 2016
& \centering Sentences
& \centering - & \centering 95.2\%
\tabularnewline
\hline

\centering Petridis et al. \cite{7472088}&  \centering DCT + LSTM
& \centering OuluVS, AVLetters & 2016
& \centering Letters
& \centering -  & \centering 81.8\%
\tabularnewline
\hline

Petridis et al. \cite{7472088}& \centering DCT + LSTM
& OuluVS, AVLetters & 2016
& \centering Sentences
& \centering -  & \centering 58.1\% 
\tabularnewline
\hline

\centering Philip et al. \cite{8356433}&  \centering DNN
& \centering XM2VTS & \centering 2017
& \centering Words
& \centering -  & \centering 86.3\% 
\tabularnewline
\hline

\centering Eric Tatulli et al. \cite{7952701}&  \centering CNN
& Custom & 2017
& \centering Sentences
& \centering -  & \centering 80.4\%
\tabularnewline
\hline

\centering P. Sindhura \cite{9001505} & \centering AlexNet, Inception V3 & MIRACL-VC1 &
2018
& \centering Words 
&  \centering 37.1\% & \centering -
\tabularnewline
\hline

\centering H. Gupta et al. \cite{8530509} & \centering CNN batch normalization & \centering MIRACL-VC1 &
2018
& \centering Words 
& \centering 96.5\% & \centering 52.9\%. 
\tabularnewline
\hline

\centering M. A. Abrar et al. \cite{9042439} & \centering VGGNet & \centering MIRACL-VC1 &
2019&
\centering Words &
\centering 94.86\% &
\centering 60 \% 
\tabularnewline 
\hline

\centering D. Parekh et al. \cite{9033664} &
\centering Autoencoders + LSTMs &
\centering MIRACL-VC1 &
\centering 2019 &
\centering Words  
& \centering 92.29\% & \centering 63.22\%
\tabularnewline  
\hline

Y.Lu et al. \cite{app9081599}&  \centering CNN + LSTM
& Custom & 2019 
& \centering Words 
& \centering 88.2\% & \centering 84.9\%
\tabularnewline
\hline

Guan et al. \cite{8929014}& \centering Fuzzy CNN
& \centering Custom & \centering 2020
& \centering Phrases
& \centering - & \centering 98.4\%
\tabularnewline
\hline

\centering N. K. Mudaliar et al. \cite{9137926}& \centering ResNet with 3D Conv. Layers + GRU
& \centering LRW & 2020
& \centering Words
& \centering 90\% & \centering 88 \%
\tabularnewline
\hline

\centering Santos et al. \cite{santos2021speaker} &
\centering Inception V3 &
\centering GRID &
\centering 2021 &
\centering Sentences &
\centering 53\% &
\centering - \tabularnewline \hline

Yang et al. \cite{9190780} & TVSR-Net
& GRID & 2021
& \centering Sentences
& \centering -  & \centering -
\tabularnewline
\hline

Qun Zhan et al. \cite{9506396}& DVSR-Net
& GRID, VSA & 2021
& \centering Sentences
& \centering - & \centering -
\tabularnewline
\hline

Xue et al. \cite{xue2023lipformer} &  Cross-Modal Fusion via Transformer 
& CMLR & 2021
& \centering Sentences
& \centering - & \centering 57\%
\tabularnewline
\hline

Ma et al. \cite{ma2022visual} & Hybrid Res-Net 
& LRW, LRS & 2022
& \centering Sentences
& \centering - & \centering 70.5\%
\tabularnewline 
\hline

Nemani et al. \cite{nemani2022deep} & 3D CNNs &
MIRACL-VC1 & 2022
& \centering Words
& \centering 71.3\% 
& \centering 70.2\%
\tabularnewline
\hline

Nemani et al. \cite{nemani2022deep} & 3D CNNs &
Custom & 2022
& \centering Words
& \centering 80.2\% & \centering 77.9\%
\tabularnewline
\hline

Kim et al. \cite{kim2023prompt} & Prompt Tuning &
GRID & 2023
& \centering Sentences 
& \centering - & \centering - 
\tabularnewline
\hline

\end{tabular}}
\end{table*}

In 2020, \textbf{Cheng Cheng Guan et al. \cite{8929014}} proposed a modern architecture fuzzy CNN model for lip image segmentation. This model integrates the deep CNN subnet and the fuzzy learning module. The integration of deep CNNs and fuzzy learning modules has been proposed as a way to improve the accuracy and interpretability of deep learning models. CNNs have been shown to be highly effective at feature extraction and pattern recognition tasks in many fields, including computer vision, speech recognition, and natural language processing. However, CNNs can be difficult to interpret, and their decisions may not always be transparent. On the other hand, fuzzy logic systems are a form of artificial intelligence that can handle imprecise or uncertain data and provide transparent, interpretable results. By integrating a fuzzy learning module with a deep CNN subnet, it is possible to combine the feature extraction capabilities of the CNN with the interpretability of fuzzy logic. The deep convolutional subnet is used for the extraction of multiscale visual features, and the fuzzy learning module is used to extract high-level semantic features while considering ambiguities and non-linearities. Additionally, a training strategy was employed to identify the ideal parameters for the CNN subnet and fuzzy module. \textbf{A customized dataset with more than 48,000 images was used to evaluate the lip segmentation using this model and a testing accuracy of 98.4\% was depicted}. \textbf{Navin et al. \cite{9137926}} proposed a deep learning model for VSR to perform word-level classification. ResNet architecture is used along with 3D convolution layers and Gated recurrent units (GRU). The entire video sequence is used as input for the architecture. On the BBC data set, the architecture obtained  an accuracy of 90\%, and on the custom video dataset, it obtained an accuracy of 88\%. The major drawback of this study is that it doesn’t perform well for subjects with facial hair. 

In 2021, \textbf{Santos et al. \cite{santos2021speaker}} proposed a study in which transfer learning was utilized to train the Inception v3 CNN model, which already had pre-trained weights from IMAGENET, with the GRID corpus. The resulting speech recognition outcomes were deemed satisfactory, with a precision of 0.61, recall of 0.53, and F1-score of 0.51. The lip-reading model demonstrated the ability to learn relevant features automatically, as evidenced by visualization, and achieved speaker-independent outcomes that were comparable to those achieved by human lip readers on the GRID corpus. The researchers identified limitations that are consistent with those encountered by humans, which could limit deep learning performance in real-world scenarios. \textbf{Soundarya et al. \cite{soundarya2021visual}} proposed a methodology that involves utilizing a combination of convolutional neural networks and Hidden Markov models, known as CNN-HMM, for lip reading. CNNs are able to identify important features within input images, making it easier to distinguish differences among the images. The HMM is then applied to handle the dynamics of the image sequence. To implement this method, the incoming video is first converted into a series of images, and then selected images are used for further analysis. The use of HMM provides a robust approach to speech recognition, resulting in a highly reliable lip reading model.

\textbf{Yang et al. \cite{9190780}} have proposed a novel DNN for improving lipreading performance in the absence of a speaker. The proposed network comprises two components, namely the Speaker Confusion Block (SC-Block) and the transformer-based VSR Network (TVSR-Net). The TVSR-Net is designed for recognizing speech and extracting lip characteristics, while the SC-Block is intended to achieve speaker normalization by reducing the impact of different talking styles and habits. The proposed system also incorporates a multi-task learning (MTL) system for network optimization. Experimental results on the GRID dataset demonstrate that the proposed network achieves excellent speaker-independent recognition performance even with limited training data. In 2021, \textbf{Qun Zhang et al. \cite{9506396}} proposed a method for enhancing the performance of VSR in the speaker-independent scenario by using disentangled feature representation and adversarial learning. The proposed disentanglement component effectively separates identity-irrelevant and content-related features from the lip image sequence to improve the VSR's accuracy. Experimental results demonstrate that the proposed method can successfully separate identification and content characteristics, and the refined content features have a high level of discrimination in speech content recognition and are independent of different speaking styles. The performance of the proposed method is evaluated using word error rate on two different datasets, GRID and VSA, which shows significant improvement over the state-of-the-art methods. In 2022, \textbf{Ma et al. \cite{ma2022visual}} suggested a VSR method that achieves state-of-the-art performance by carefully developing a model and using larger datasets. They also stressed hyperparameter tweaking, which may improve present designs, and time-masking, which makes the network focus more on context. The authors used six datasets and found that larger datasets improve performance, consistent with existing findings. The authors conducted an ablation study and have a word error rate of  \textbf{29.5±0.4 on LR2 and  35.8±0.5 on the LR3 datasets}. The proposed VSR algorithm surpasses all others trained on publicly available English, Spanish, and Mandarin datasets.

\textbf{Xue et al. \cite{xue2023lipformer}}  suggested a cross-modal Transformer framework for sentence-level lipreading that can generalize to unseen speakers by calibrating visual changes using landmarks as motion trajectories. Cross-attention to exhibit cross-modal fusion may help the model align features. Extensive experimentation was conducted on the CMLR dataset and the sentence level accuracy achieved was 57\%. The approach generalizes to unseen speakers, according to considerable experimental evidence. The recommended solution outperforms five cutting-edge algorithms for unseen and overlapping speakers, including those of different ethnicities. \textbf{Nemani et al. \cite{nemani2022deep}} pointed out that the current state of Video-based Speech Recognition (VSR) systems faces various challenges, such as insufficient training data, lack of holistic feature representation, and lower accuracy rates. To overcome these limitations, a novel and scalable VSR system is proposed in this study. The system uses the user's videotape to determine the spoken word, and a customized 3-Dimensional Convolutional Neural Network (3D CNN) architecture is proposed for feature extraction and mapping prediction probabilities. To validate the concept of person independence, a customized dataset is created similar to the MIRACL-VC1 dataset. The proposed system is robust to various lighting conditions across multiple devices and achieves a \textbf{training accuracy of 80.2\% and testing accuracy of 77.9\%} in predicting the spoken word by the user. Also, the authors imply the presence of an end-to-end solution for isolated word-level VSR by the introduction of an \textbf{edge device prototype} to predict the word spoken by the user. The research conducted by \textbf{Kim et al. \cite{kim2023prompt}} in 2023 presents findings on the impact of prompt adjustment in speaker-adaptive VSR models. In contrast to previous studies, the authors utilized target speaker adaptation data instead of model parameters, taking advantage of advancements in NLP. They explored various prompt adjustment methods, including the addition, padding, and concatenation of CNN Transformer VSR models. However, the Transformer variation architecture limited quick tweaking methods. The research revealed that even though the pre-trained VSR model should be developed with a large speaker variation, a small amount of adaptation data, such as less than 5 minutes, could improve the model's performance on unknown speakers with quick tuning. The authors also conducted a comparison of prompt performance and parameters for fine-tuning. The study's findings indicate that prompt adjustment using target speaker adaptation data can significantly improve unseen speaker performance in VSR models. These results have implications for the development of speaker-adaptive VSR models and could inform future research in the field.  

\section{Key Findings and Comparative Analysis}
After describing the different methodologies used for speaker-independent VSR, in this section, we intend to compare the above methodologies based on their accuracy, scalability, feature representation, and deployment. For any VSR system, the above factors are considered essential to measure its performance and robustness to various conditions.

\begin{table*}[htbp]

	\centering
	\caption{Attribute Comparison of the datasets for speaker independent VSR}
 \label{Tab: datasetsa}
	\resizebox{\linewidth}{!}{\begin{tabular}{|c|c|c|c|c|c|c|c|c|}
	\hline
 
 \centering \textbf{Dataset} & 
 \centering \textbf{Year} & 
 \centering \textbf{Lip Region Features} & 
 \centering \textbf{Facial Features} & 
 \centering \textbf{Laboratory Conditions} & 
 \centering \textbf{Isolated} & 
 \centering \textbf{Magnitude} & 
 \centering \textbf{Open Source} & 
 \centering \textbf{HD Resolution} \tabularnewline 
 \hline

\centering Tulips1 \cite{movellan1996channel} &
\centering 1996 &
\centering $\textcolor{green}{\checkmark}$ &
\centering $\textcolor{red}{\times}$ &
\centering $\textcolor{green}{\checkmark}$ &
\centering $\textcolor{green}{\checkmark}$ &
\centering $\textcolor{red}{\times}$ & 
\centering $\textcolor{green}{\checkmark}$ & 
\centering $\textcolor{red}{\times}$ \tabularnewline
\hline

\centering M2VTS \cite{kittler1997combining} &
\centering 1997 &
\centering $\textcolor{green}{\checkmark}$ &
\centering $\textcolor{green}{\checkmark}$ &
\centering $\textcolor{green}{\checkmark}$ &
\centering $\textcolor{green}{\checkmark}$ &
\centering $\textcolor{red}{\times}$ &
\centering $\textcolor{red}{\times}$ & 
\centering $\textcolor{red}{\times}$ \tabularnewline
\hline

\centering AVLetters \cite{982900} &
\centering 1998 &
\centering $\textcolor{green}{\checkmark}$ &
\centering $\textcolor{red}{\times}$ &
\centering $\textcolor{green}{\checkmark}$ &
\centering $\textcolor{green}{\checkmark}$ &
\centering $\textcolor{red}{\times}$ &
\centering $\textcolor{green}{\checkmark}$ & 
\centering $\textcolor{red}{\times}$ \tabularnewline
\hline

\centering AVLetters2 \cite{inproceedings} &
\centering 2008 &
\centering $\textcolor{green}{\checkmark}$ &
\centering $\textcolor{red}{\times}$ &
\centering $\textcolor{green}{\checkmark}$ &
\centering $\textcolor{green}{\checkmark}$ &
\centering $\textcolor{red}{\times}$ &
\centering $\textcolor{green}{\checkmark}$ & 
\centering $\textcolor{green}{\checkmark}$ \tabularnewline
\hline

\centering MIRACL-VC1 \cite{rekik2014new} & 
\centering 2011 & 
\centering $\textcolor{green}{\checkmark}$ &
\centering $\textcolor{green}{\checkmark}$ &
\centering $\textcolor{green}{\checkmark}$ &
\centering $\textcolor{green}{\checkmark}$ &
\centering $\textcolor{red}{\times}$ &
\centering $\textcolor{green}{\checkmark}$ &
\centering $\textcolor{red}{\times}$ \tabularnewline
\hline

\centering OuluVS \cite{5208233} & 
\centering 2015 & 
\centering $\textcolor{green}{\checkmark}$ &
\centering $\textcolor{green}{\checkmark}$ &
\centering $\textcolor{green}{\checkmark}$ &
\centering $\textcolor{green}{\checkmark}$ &
\centering $\textcolor{red}{\times}$ &
\centering $\textcolor{red}{\times}$ &
\centering $\textcolor{red}{\times}$ \tabularnewline
\hline

\centering LRW \cite{Chung16} &
\centering 2016 &
\centering $\textcolor{green}{\checkmark}$ &
\centering $\textcolor{green}{\checkmark}$ &
\centering $\textcolor{red}{\times}$ &
\centering $\textcolor{green}{\checkmark}$ &
\centering $\textcolor{green}{\checkmark}$ &
\centering $\textcolor{green}{\checkmark}$ &
\centering $\textcolor{red}{\times}$ \tabularnewline
\hline

\centering LRS2 \cite{Chung17} &
\centering 2017 &
\centering $\textcolor{green}{\checkmark}$ &
\centering $\textcolor{green}{\checkmark}$ &
\centering $\textcolor{red}{\times}$ &
\centering $\textcolor{green}{\checkmark}$ &
\centering $\textcolor{green}{\checkmark}$ &
\centering $\textcolor{green}{\checkmark}$ &
\centering $\textcolor{red}{\times}$ \tabularnewline
\hline

\centering LRS3-TED \cite{DBLP:journals/corr/abs-1809-00496} &
\centering 2017 &
\centering $\textcolor{green}{\checkmark}$ &
\centering $\textcolor{green}{\checkmark}$ &
\centering $\textcolor{red}{\times}$ &
\centering $\textcolor{green}{\checkmark}$ &
\centering $\textcolor{green}{\checkmark}$ &
\centering $\textcolor{green}{\checkmark}$ &
\centering $\textcolor{red}{\times}$ \tabularnewline
\hline

\centering AVDigits \cite{7780758} &
\centering 2018 &
\centering $\textcolor{green}{\checkmark}$ &
\centering $\textcolor{green}{\checkmark}$ &
\centering $\textcolor{red}{\times}$ &
\centering $\textcolor{green}{\checkmark}$ &
\centering $\textcolor{red}{\times}$ &
\centering $\textcolor{green}{\checkmark}$ &
\centering $\textcolor{red}{\times}$ \tabularnewline
\hline

\centering GLips \cite{schwiebert2022multimodal} &
\centering 2022 &
\centering $\textcolor{green}{\checkmark}$ &
\centering $\textcolor{red}{\times}$ & 
\centering $\textcolor{red}{\times}$ &
\centering $\textcolor{red}{\times}$ &
\centering $\textcolor{green}{\checkmark}$ &
\centering $\textcolor{green}{\checkmark}$ &
\centering $\textcolor{red}{\times}$  \tabularnewline
\hline

\centering CN-CVS/Speech \cite{chencn} &
\centering 2022 &
\centering $\textcolor{green}{\checkmark}$ &
\centering $\textcolor{green}{\checkmark}$ &
\centering $\textcolor{red}{\times}$ &
\centering $\textcolor{red}{\times}$ &
\centering $\textcolor{green}{\checkmark}$ &
\centering $\textcolor{green}{\checkmark}$ &
\centering $\textcolor{red}{\times}$  \tabularnewline
\hline

\centering CN-CVS/News \cite{chencn} &
\centering 2022 &
\centering $\textcolor{green}{\checkmark}$ &
\centering $\textcolor{green}{\checkmark}$ &
\centering $\textcolor{red}{\times}$ &
\centering $\textcolor{red}{\times}$ &
\centering $\textcolor{green}{\checkmark}$ &
\centering $\textcolor{green}{\checkmark}$ &
\centering $\textcolor{red}{\times}$  \tabularnewline
\hline

\centering RUSAVIC \cite{ivanko-etal-2022-rusavic} &
\centering 2022 &
\centering $\textcolor{green}{\checkmark}$ &
\centering $\textcolor{green}{\checkmark}$ &
\centering $\textcolor{red}{\times}$ &
\centering $\textcolor{green}{\checkmark}$ &
\centering $\textcolor{green}{\checkmark}$ &
\centering $\textcolor{green}{\checkmark}$ &
\centering $\textcolor{green}{\checkmark}$  \tabularnewline
\hline

\centering OLKAVS \cite{park2023olkavs} &
\centering 2023 &
\centering $\textcolor{green}{\checkmark}$ &
\centering $\textcolor{green}{\checkmark}$ &
\centering $\textcolor{green}{\checkmark}$ &
\centering $\textcolor{green}{\checkmark}$ &
\centering $\textcolor{green}{\checkmark}$ &
\centering $\textcolor{red}{\times}$ &
\centering $\textcolor{green}{\checkmark}$ \tabularnewline
\hline

	\end{tabular}}
\end{table*}

\subsection{Comparing Datasets: Key Findings and Attributes Analysis}
In this section, a comparative analysis is presented for different datasets on seven attributes: the presence of \textbf{lip features, availability of facial features, recording conditions in a laboratory environment, isolation of person in the recording, dataset size, open-source availability,} and \textbf{dataset resolution} in Table \ref{Tab: datasetsa}. These attributes were chosen as key factors to examine and differentiate the datasets. Based on the table, it can be observed that all of the datasets exhibit the presence of lip features, which are discernible through the movements of the lips during the pronunciation of words or sentences. Nevertheless, there are some datasets that do not prioritize the inclusion of facial features or the complete faces of the individuals being recorded \cite{movellan1996channel, 982900, inproceedings, schwiebert2022multimodal}. Although the lip region is a crucial area for speech perception and recognition as it conveys significant information about phonemes and syllables, facial features also play a crucial role. While lip movements can convey a significant amount of phonetic information, other facial features such as eyebrows, eyes, and cheeks also contribute to speech perception. For instance, raising the eyebrows can indicate surprise or emphasis, while the narrowing of the eyes can convey anger or frustration. These facial features and lip movements provide a more comprehensive view of the speaker's speech, making it easier to recognize and understand. Therefore, it is essential to include both lip features and other facial features when preparing a VSR dataset to improve recognition accuracy. For speaker independent VSR, preparing the dataset with a single person present in each frame is crucial. This ensures that the VSR system can accurately recognize the speech signal independent of the speaker's identity. By \textbf{isolating} the individual in each frame, the VSR model can learn to recognize and generalize speech patterns more effectively. Additionally, this approach reduces the variability in the dataset caused by differences in the appearance and speech patterns of multiple individuals, leading to a more robust and reliable recognition system. The table shows that the datasets cited as \cite{schwiebert2022multimodal, chencn} deviate from the standard practice of isolating a single individual in each frame. These datasets contain multiple persons in each frame, so they are not isolated. 

\textbf{Laboratory conditions} are necessary for speaker-independent VSR as they provide a controlled environment for data collection. In a laboratory, lighting, camera position, and background can be standardized, reducing the dataset's variability and ensuring that the visual cues for speech recognition are consistent across all samples. Moreover, a laboratory environment also allows for the use of high-quality equipment, such as cameras and microphones, which can capture speech signals with higher accuracy and resolution. In contrast, if the data is collected outside of the laboratory, there can be variations in lighting, background noise, and camera position, which can affect the quality of the data and make it more challenging to recognize speech accurately. Datasets such as LRW and CN-CVS were not recorded under laboratory conditions, as they were captured from television shows. As a result, the data in these datasets may be more diverse and complex than data captured under controlled laboratory conditions. It is true that datasets recorded in a laboratory tend to be smaller in size than those captured in natural environments, which is illustrated by the LRW, LRS2, and CN-CVS datasets. This is because laboratory conditions limit the variability of the data, resulting in a smaller range of speech patterns and visual cues for recognition. In contrast, datasets captured in natural environments may contain a larger amount of data due to the variability in speech patterns and visual cues that occur in different settings. However, this larger dataset size comes at the cost of increased complexity and variability, making it more challenging to build an accurate and reliable VSR model. Therefore, when preparing a VSR dataset, the trade-off between dataset size and variability must be considered. While larger datasets may provide more information for recognition, smaller datasets recorded in a laboratory can be more controlled and reliable, leading to better recognition accuracy in some cases. 

\begin{table*}[ht]
	\centering
\caption{Attribute comparison of VSR works from the 90s: A Tabulated Overview}
	\label{Tab:Comparison1990}
	\resizebox{\linewidth}{!}{
	\begin{tabular}{|c|p{20mm}|p{15mm}|p{20mm}|p{15mm}|p{20mm}|p{15mm}|}
	\hline
	\centering \textbf{Existing Works}& 
	\centering \textbf{Holistic Feature Representation}& 
	\centering \textbf{Speaker Independence}& 
	\centering \textbf{Edge Device Deployment}& 
	\centering \textbf{Real-time Testing}&
    \centering \textbf{Benchmark Dataset Evaluation}&
    \centering \textbf{Accuracy $\geq$ 80\%}
\tabularnewline
\hline

\centering Gelder et al. \cite{gelder1991face} &
\centering $\textcolor{red}{\times}$ &
\centering $\textcolor{red}{\times}$ &
\centering $\textcolor{red}{\times}$ &
\centering $\textcolor{green}{\checkmark}$ &
\centering $\textcolor{red}{\times}$ &
\centering $\textcolor{red}{\times}$ \tabularnewline
\hline

\centering Walden et al. \cite{walden1993benefit} &
\centering $\textcolor{red}{\times}$ &
\centering $\textcolor{red}{\times}$ &
\centering $\textcolor{red}{\times}$ &
\centering $\textcolor{green}{\checkmark}$ &
\centering $\textcolor{red}{\times}$ &
\centering $\textcolor{red}{\times}$  \tabularnewline
\hline

\centering Silsbee et al. \cite{silsbee1993audio} & 
\centering $\textcolor{red}{\times}$ &
\centering $\textcolor{red}{\times}$ &
\centering $\textcolor{red}{\times}$ &
\centering $\textcolor{green}{\checkmark}$ &
\centering $\textcolor{red}{\times}$ &
\centering $\textcolor{red}{\times}$  \tabularnewline
\hline

\centering Javier et al. \cite{movellan1994visual} &
\centering $\textcolor{green}{\checkmark}$ &
\centering $\textcolor{red}{\times}$ &
\centering $\textcolor{red}{\times}$ &
\centering $\textcolor{green}{\checkmark}$ &
\centering $\textcolor{red}{\times}$ &
\centering $\textcolor{red}{\times}$  \tabularnewline
\hline

\centering Marassa et al. \cite{marassa1995visual} &
\centering $\textcolor{red}{\times}$ &
\centering $\textcolor{red}{\times}$ &
\centering $\textcolor{red}{\times}$ &
\centering $\textcolor{green}{\checkmark}$ &
\centering $\textcolor{red}{\times}$ &
\centering $\textcolor{red}{\times}$  \tabularnewline
\hline

\centering Bregler et al. \cite{bregler1995nonlinear} &
\centering $\textcolor{green}{\checkmark}$ &
\centering $\textcolor{green}{\checkmark}$ &
\centering $\textcolor{red}{\times}$ &
\centering $\textcolor{green}{\checkmark}$ &
\centering $\textcolor{red}{\times}$ &
\centering $\textcolor{red}{\times}$ \tabularnewline
\hline

\centering Luettin et al. \cite{luettin1996statistical} &
\centering $\textcolor{green}{\checkmark}$ &
\centering $\textcolor{red}{\times}$ &
\centering $\textcolor{red}{\times}$ &
\centering $\textcolor{red}{\times}$ &
\centering $\textcolor{green}{\checkmark}$ &
\centering $\textcolor{green}{\checkmark}$ \tabularnewline
\hline

\centering Luettin et al. \cite{luettin1996visual} &
\centering $\textcolor{green}{\checkmark}$ &
\centering $\textcolor{red}{\times}$ &
\centering $\textcolor{red}{\times}$ &
\centering $\textcolor{red}{\times}$ &
\centering $\textcolor{green}{\checkmark}$ &
\centering $\textcolor{green}{\checkmark}$ \tabularnewline
\hline

\centering Chiou et al. \cite{chiou1997lipreading} &
\centering $\textcolor{red}{\times}$ &
\centering $\textcolor{green}{\checkmark}$ &
\centering $\textcolor{red}{\times}$ &
\centering $\textcolor{green}{\checkmark}$ &
\centering $\textcolor{red}{\times}$ &
\centering $\textcolor{red}{\times}$ \tabularnewline
\hline

\centering Yu et al. \cite{yu1997lipreading} &
\centering $\textcolor{red}{\times}$ &
\centering $\textcolor{red}{\times}$ &
\centering $\textcolor{red}{\times}$ &
\centering $\textcolor{red}{\times}$ &
\centering $\textcolor{red}{\times}$ &
\centering $\textcolor{red}{\times}$ \tabularnewline
\hline

\centering G. Rabi et al. \cite{614788} & 
\centering $\textcolor{green}{\checkmark}$ &
\centering $\textcolor{green}{\checkmark}$ &
\centering $\textcolor{red}{\times}$ &
\centering $\textcolor{green}{\checkmark}$ &
\centering $\textcolor{red}{\times}$ &
\centering $\textcolor{red}{\times}$ \tabularnewline
\hline

\centering Nanaku et al. \cite{nankaku1999intensity} &
\centering $\textcolor{green}{\checkmark}$ &
\centering $\textcolor{red}{\times}$ &
\centering $\textcolor{red}{\times}$ &
\centering $\textcolor{red}{\times}$ &
\centering $\textcolor{green}{\checkmark}$ &
\centering $\textcolor{red}{\times}$ \tabularnewline
\hline

\centering Javier et al. \cite{movellan1999diffusion} &
\centering $\textcolor{green}{\checkmark}$ &
\centering $\textcolor{green}{\checkmark}$ &
\centering $\textcolor{red}{\times}$ &
\centering $\textcolor{green}{\checkmark}$ &
\centering $\textcolor{green}{\checkmark}$ &
\centering $\textcolor{red}{\times}$ \tabularnewline
\hline

\centering Baldwin et al. \cite{baldwin1999automatic} &
\centering $\textcolor{green}{\checkmark}$ &
\centering $\textcolor{red}{\times}$ &
\centering $\textcolor{red}{\times}$ &
\centering $\textcolor{green}{\checkmark}$ &
\centering $\textcolor{green}{\checkmark}$ &
\centering $\textcolor{red}{\times}$ \tabularnewline
\hline

\end{tabular}}\\
\end{table*}

It is evident that the magnitude of datasets increased drastically as the years progressed. \textbf{Resolution} is one of the most critical factors for speaker-independent VSR dataset creation because it directly affects the quality of the visual cues captured in the dataset. Visual cues for speech recognition are subtle and can be affected by even minor variations in lighting, camera position, and resolution. Higher-resolution images can capture more detailed visual cues for speech recognition, making it easier for VSR models to accurately identify and differentiate between different speech sounds. In contrast, low-resolution images may not capture enough visual information, making it more challenging for VSR models to recognize speech accurately. Moreover, higher-resolution images also provide more precise and accurate data for training and testing VSR models. This can help improve the overall accuracy and robustness of the VSR model, particularly in complex and dynamic environments. Therefore, the resolution is a critical factor in speaker-independent VSR dataset creation, and the use of high-resolution images can lead to better speech recognition performance in VSR models. From the table, it is depicted that only AVLetters2 and OLKAVIS Datasets have samples in high resolution. Based on the preceding inference, it can be inferred that the \textbf{OLKAVIS dataset} fulfills all the aforementioned requirements and is thus deemed to be an optimal dataset for the speaker-independent VSR task. It is also considered highly essential to compare the highest accuracies achieved by each dataset in the task of speaker independent VSR. Comparing the highest accuracies also facilitates evaluating and selecting appropriate datasets for specific research or application requirements. Researchers can choose datasets that have consistently shown high accuracies, indicating their suitability for training and testing speaker-independent VSR models. From Table \ref{Tab: datasetsacc}, it can be inferred that the \textbf{MIRACL-VC1 dataset}, as reported by \textbf{D. Parekh et al. \cite{9033664}}, achieved a remarkable highest accuracy of 92.29\%. On the other hand, the AVLetters2 dataset and LRS3-TED dataset, and the newly proposed multilingual datasets do not have the specific highest accuracy values mentioned in the table. This could be due to the unavailability of accurate reported results or the absence of relevant research papers. 

\begin{table}[htbp]

	\centering
	\caption{Accuracy Metrics of Speaker Independent VSR Datasets}
 \label{Tab: datasetsacc}
	\resizebox{\linewidth}{!}{\begin{tabular}{|c|c|c|}
	\hline
 
 \centering \textbf{Dataset} & 
 \centering \textbf{Highest Accuracy Achieved} & 
 \centering \textbf{Author} \tabularnewline 
 \hline

\centering Tulips1 \cite{movellan1996channel} &
\centering 74.5\% &
\centering Vanegas et al. \cite{vanegas2000lip} \tabularnewline
\hline

\centering M2VTS \cite{kittler1997combining} &
\centering 60.2\% &
\centering Luettin et al. \cite{luettin1997towards} \tabularnewline
\hline

\centering AVLetters \cite{982900} &
\centering 81.8\% &
\centering Petridis et al. \cite{7472088} \tabularnewline
\hline

\centering AVLetters2 \cite{inproceedings} &
\centering - &
\centering - \tabularnewline
\hline

\centering MIRACL-VC1 \cite{rekik2014new} & 
\centering 92.29\% & 
\centering D. Parekh et al. \cite{9033664} \tabularnewline
\hline

\centering OuluVS \cite{5208233} & 
\centering 60.2\% &
\centering Luettin et al. \cite{luettin1997towards} \tabularnewline
\hline

\centering LRW \cite{Chung16} &
\centering 70.5\% &
\centering Ma et al. \cite{ma2022visual} \tabularnewline
\hline

\centering LRS2 \cite{Chung17} &
\centering 70.5\% &
\centering Ma et al. \cite{ma2022visual} \tabularnewline
\hline

\centering LRS3-TED \cite{DBLP:journals/corr/abs-1809-00496} &
\centering - &
\centering - \tabularnewline
\hline

	\end{tabular}}
\end{table}

\subsection{Attribute Comparison in VSR: Overview from the 90s}
Table \ref{Tab:Comparison1990} clearly demonstrates the explosive growth of research in VSR during the 1990s. Seminal works in the field were evaluated on various parameters, including Holistic Feature Representation, Deployment, Testing, Dataset Evaluation, Accuracy, and speaker independence. \textbf{Holistic feature representation} \cite{sun2009action, zhou2011computationally, tanaka2003holistic} refers to the approach used in VSR that considers the whole facial area rather than just specific facial features. This involves analyzing the changes in facial features, such as the lips, tongue, and jaw, to extract information about the spoken words. The idea is to capture the overall motion of the facial features, which can provide more information about the spoken words than just analyzing individual features. Holistic feature representation is a popular approach in VSR, as it can lead to better accuracy in recognizing spoken words. Notably, the majority of works during this period focused on speaker-dependent VSR and aimed to create a basic framework that could achieve the highest possible accuracy. As a result, end-to-end solutions that satisfied all parameters were not yet available. The table also suggests that Holistic Feature Representation emerged as a significant aspect of VSR research in the late 90s, with many works displaying this attribute. Moreover, since the majority of datasets used in these studies were customized and recorded under laboratory conditions, it can be inferred that most works underwent real-time testing. The introduction of benchmark datasets like Tulips1 and M2VTS also made dataset evaluation an essential component of VSR research. Despite these advancements, techniques during this time were not highly efficient, with few studies achieving accuracy greater than 80\%. Additionally, speaker independence, which is a critical component of VSR, was only evident in a limited number of works.

\subsection{Exploring the Evolution of Visual Speech Recognition: An in-depth Analysis of Attribute Comparison in VSR Research from 2000 to 2010}
The period between 2000 and 2010 witnessed significant advancement in various areas of VSR. One of the noticeable improvements during this period was the remarkable enhancement in the accuracy of VSR techniques. Moreover, there was a notable emphasis on creating benchmark datasets to evaluate the performance of different VSR methodologies, which is crucial for ensuring the reliability and reproducibility of results. The table presented in this context, Table \ref{Tab:Comparison2000}, highlights the progress in this area during this period. Additionally, the decade of 2000-2010 marked the emergence of innovative ideas on deploying VSR on edge devices, which has become a crucial research area in recent times. Traditional approaches like Hidden Markov Models (HMMs), Support Vector Machines (SVMs), AdaBoost classifiers, Discrete Cosine Transform (DCT), and Contextual modeling were extensively used during this decade, leading to high accuracy with minimal error rates. The techniques were evaluated using various corpuses, including words, sentences, alphabets, and phrases, resulting in an average recognition rate of 70\% in the early 2000s and an increase to about 85\% in the late 2000s. The average error rate during this period was 10\%. However, speaker independence remained a challenging problem, as illustrated in Table \ref{Tab:Comparison2000}. Overall, the decade of 2000-2010 marked significant progress in the field of VSR by enhancing the accuracy of existing techniques, creating benchmark datasets, and exploring innovative ideas for deploying VSR on edge devices.

\subsection{Analyzing the Evolution of Visual Speech Recognition: A Comprehensive Study of Attribute Comparison in VSR Research from 2011 to 2023}
In the current era of technological advancements, VSR has seen substantial growth in various aspects since 2011. The research has been focused on exploring new techniques and architectures for VSR to enhance its performance, and benchmark datasets have become an integral part of the research which can be depicted in Table \ref{Tab:Comparison2011}. \textbf{Benchmark dataset evaluation} is an essential aspect of VSR because it allows for the objective evaluation of different methodologies. Using benchmark datasets, researchers can compare the performance of different VSR techniques using the same data, allowing for fair and meaningful comparisons. Moreover, benchmark datasets help in understanding the limitations of the existing VSR systems and provide a basis for further development. These datasets are designed to cover a wide range of challenging conditions, such as variations in lighting, background, speaker demographics, and camera angles. They also provide a standardized testing protocol, which is crucial for comparing the performance of different systems. Using benchmark datasets, researchers can identify the strengths and weaknesses of different VSR techniques and provide a standardized metric to evaluate the performance of new techniques. Additionally, benchmark datasets can be used to train machine learning algorithms, which can learn to recognize speech patterns and improve the overall accuracy of the VSR system. The majority of the datasets mentioned in Table \ref{Tab: datasetsa} were formulated in this period The introduction of deep learning architectures like CNNs, RNNs, and LSTM, among others, has had a significant impact on improving the accuracy of VSR. In addition, recent works have emphasized the importance of developing speaker-independent models, which are essential for VSR applications in real-world scenarios. Based on the analysis of the table, it is evident that speaker independence has been a crucial aspect addressed in all the VSR works in the last decade. A majority of the methodologies have also incorporated the property of holistic feature representation. Furthermore, there has been a remarkable improvement in the accuracy of the proposed solutions in the late 2010s, making them the current state-of-the-art.

\begin{table*}[ht]
	\centering
\caption{Attribute comparison of VSR works from 2000-2010: A Tabulated Overview.}
	\label{Tab:Comparison2000}
	\resizebox{\linewidth}{!}{
	\begin{tabular}{|c|p{20mm}|p{15mm}|p{20mm}|p{15mm}|p{20mm}|p{15mm}|}
	\hline
	\centering \textbf{Existing Works}& 
	\centering \textbf{Holistic Feature Representation}& 
	\centering \textbf{Speaker Independence}& 
	\centering \textbf{Edge Device Deployment}& 
	\centering \textbf{Real-time Testing}&
    \centering \textbf{Benchmark Dataset Evaluation}&
    \centering \textbf{Accuracy $\geq$ 80\%}
\tabularnewline
\hline

\centering Vanegas et al. \cite{vanegas2000lip} & 
\centering $\textcolor{green}{\checkmark}$ &
\centering $\textcolor{red}{\times}$ &
\centering $\textcolor{red}{\times}$ &
\centering $\textcolor{red}{\times}$ &
\centering $\textcolor{green}{\checkmark}$ &
\centering $\textcolor{red}{\times}$ \tabularnewline
\hline

\centering Nankaku et al. \cite{nankaku2000normalized} &
\centering $\textcolor{red}{\times}$ &
\centering $\textcolor{red}{\times}$ &
\centering $\textcolor{red}{\times}$ &
\centering $\textcolor{red}{\times}$ &
\centering $\textcolor{green}{\checkmark}$ &
\centering $\textcolor{red}{\times}$ \tabularnewline
\hline

\centering Gordan et al. \cite{gordan2002support} & 
\centering $\textcolor{red}{\times}$ &
\centering $\textcolor{red}{\times}$ &
\centering $\textcolor{red}{\times}$ &
\centering $\textcolor{red}{\times}$ &
\centering $\textcolor{green}{\checkmark}$ &
\centering $\textcolor{red}{\times}$ \tabularnewline
\hline

\centering Foo et al. \cite{1202350} & 
\centering $\textcolor{red}{\times}$ &
\centering $\textcolor{red}{\times}$ &
\centering $\textcolor{red}{\times}$ &
\centering $\textcolor{green}{\checkmark}$ & 
\centering $\textcolor{red}{\times}$ &
\centering $\textcolor{red}{\times}$ \tabularnewline
\hline

\centering Yao et al. \cite{yao2003visual} & 
\centering $\textcolor{green}{\checkmark}$ & 
\centering $\textcolor{red}{\times}$ &
\centering $\textcolor{red}{\times}$ &
\centering $\textcolor{red}{\times}$ &
\centering $\textcolor{green}{\checkmark}$ & 
\centering $\textcolor{red}{\times}$ \tabularnewline
\hline

\centering  Anwar et al. \cite{anwar2004learning} &
\centering $\textcolor{red}{\times}$ &
\centering $\textcolor{red}{\times}$ &
\centering $\textcolor{red}{\times}$ &
\centering $\textcolor{red}{\times}$ &
\centering $\textcolor{green}{\checkmark}$ & 
\centering $\textcolor{red}{\times}$ \tabularnewline
\hline

\centering Foo et al. \cite{foo2004recognition} &
\centering $\textcolor{red}{\times}$ &
\centering $\textcolor{red}{\times}$ &
\centering $\textcolor{red}{\times}$ &
\centering $\textcolor{green}{\checkmark}$ & 
\centering $\textcolor{red}{\times}$ &
\centering $\textcolor{green}{\checkmark}$ \tabularnewline
\hline

\centering Saenko et al. \cite{1544886} &
\centering $\textcolor{red}{\times}$ &
\centering $\textcolor{red}{\times}$ &
\centering $\textcolor{red}{\times}$ &
\centering $\textcolor{red}{\times}$ &
\centering $\textcolor{green}{\checkmark}$ & 
\centering $\textcolor{red}{\times}$ \tabularnewline
\hline

\centering Sagheer et al. \cite{1577194} &
\centering $\textcolor{red}{\times}$ &
\centering $\textcolor{green}{\checkmark}$ & 
\centering $\textcolor{red}{\times}$ &
\centering $\textcolor{red}{\times}$ &
\centering $\textcolor{green}{\checkmark}$ &
\centering $\textcolor{red}{\times}$ \tabularnewline
\hline

\centering Lee et al. \cite{lee2006training} & 
\centering $\textcolor{green}{\checkmark}$ & 
\centering $\textcolor{red}{\times}$ &
\centering $\textcolor{red}{\times}$ &
\centering $\textcolor{red}{\times}$ &
\centering $\textcolor{green}{\checkmark}$ &
\centering $\textcolor{red}{\times}$ \tabularnewline
\hline

\centering Yau et al. \cite{1663790} & 
\centering $\textcolor{red}{\times}$ &
\centering $\textcolor{red}{\times}$ &
\centering $\textcolor{red}{\times}$ &
\centering $\textcolor{green}{\checkmark}$ &
\centering $\textcolor{red}{\times}$ &
\centering $\textcolor{green}{\checkmark}$ \tabularnewline
\hline

\centering Leung et al. \cite{1699666} &
\centering $\textcolor{red}{\times}$ &
\centering $\textcolor{green}{\checkmark}$ &
\centering $\textcolor{red}{\times}$ &
\centering $\textcolor{green}{\checkmark}$ &
\centering $\textcolor{red}{\times}$ &
\centering $\textcolor{green}{\checkmark}$ \tabularnewline
\hline

\centering Yu et al. \cite{yu2007new} &
\centering $\textcolor{red}{\times}$ &
\centering $\textcolor{red}{\times}$ &
\centering $\textcolor{red}{\times}$ &
\centering $\textcolor{green}{\checkmark}$ &
\centering $\textcolor{red}{\times}$ &
\centering $\textcolor{green}{\checkmark}$ \tabularnewline
\hline

\centering Jie et al. \cite{4712008} &
\centering $\textcolor{green}{\checkmark}$ &
\centering $\textcolor{green}{\checkmark}$ &
\centering $\textcolor{red}{\times}$ &
\centering $\textcolor{red}{\times}$ &
\centering $\textcolor{green}{\checkmark}$ &
\centering $\textcolor{green}{\checkmark}$ \tabularnewline
\hline

\centering Wang et al. \cite{4667054} &
\centering $\textcolor{red}{\times}$ &
\centering $\textcolor{red}{\times}$ &
\centering $\textcolor{red}{\times}$ &
\centering $\textcolor{green}{\checkmark}$ &
\centering $\textcolor{red}{\times}$ &
\centering $\textcolor{green}{\checkmark}$ \tabularnewline
\hline

\centering Rajavel et al. \cite{rajavel2009static} &
\centering $\textcolor{green}{\checkmark}$ &
\centering $\textcolor{red}{\times}$ &
\centering $\textcolor{red}{\times}$ &
\centering $\textcolor{green}{\checkmark}$ &
\centering $\textcolor{red}{\times}$ &
\centering $\textcolor{green}{\checkmark}$ \tabularnewline
\hline

\centering Pass et al. \cite{5652630} & 
\centering $\textcolor{red}{\times}$ &
\centering $\textcolor{red}{\times}$ &
\centering $\textcolor{red}{\times}$ &
\centering $\textcolor{red}{\times}$ &
\centering $\textcolor{green}{\checkmark}$ &
\centering $\textcolor{red}{\times}$  \tabularnewline
\hline

\centering Lu et al. \cite{5713068} &
\centering $\textcolor{green}{\checkmark}$ &
\centering $\textcolor{red}{\times}$ &
\centering $\textcolor{red}{\times}$ &
\centering $\textcolor{green}{\checkmark}$ &
\centering $\textcolor{red}{\times}$ &
\centering $\textcolor{green}{\checkmark}$ \tabularnewline
\hline

\end{tabular}}\\
\end{table*}

\subsection{The Vital Combination of DL Architecture and Proper Datasets}
The combination of DL architecture and a proper dataset is crucial for the success of speaker-independent VSR systems. DL architectures are designed to learn and extract relevant features from the data, allowing them to automatically adapt to the variability in speech signals caused by factors such as speaker identity, accent, and speaking style. These architectures are capable of modeling complex relationships between input features and output labels, which is essential for accurate speech recognition. However, the performance of DL models heavily depends on the quality and size of the training dataset. A proper dataset must be diverse and representative of the target population, containing a wide range of speakers, accents, and speaking styles. The dataset should also be annotated with accurate transcription or labeling, allowing the DL model to learn the relationship between acoustic features and linguistic content. Without a proper dataset, the DL model may not be able to capture the full range of variability in speech signals, leading to poor performance on unseen speakers or in noisy environments. Similarly, without an appropriate DL architecture, the model may not be able to learn the complex relationships between acoustic features and linguistic content, resulting in inaccurate speech recognition. Custom datasets are often used in VSR models to train the model to recognize a particular domain or application's specific speech patterns and vocabulary. However, if the custom dataset is not diverse or representative enough, the VSR model may not be able to generalize well to speakers with different accents or speaking styles, leading to less accuracy. While custom datasets can improve the accuracy of VSR models in specific contexts, they may also lead to less accuracy if they are not designed and implemented correctly \cite{DBLP:journals/corr/abs-1301-4558}. Traditional methods like \cite{4712008, 4667054, 6179154} have shown a significant accuracy in word-level VSR. However, they can also lead to improper generalization when used exclusively. Apart from improper datasets and traditional methodologies, a proper combination of DL models and datasets is also deemed essential. This can be shown in \cite{9001505, 8530509, 9042439, 9033664} where there is a significant difference in the Training and Test Accuracy of the model. Accurate recognition of spoken words enhances the system's user experience, reliability, and productivity, making it an essential consideration when designing and evaluating VSR systems. Hence, the right combination of DL architectures and datasets greatly improves the VSR system, which is visible in \cite{9137926, app9081599}. 

\begin{table*}[ht]
	\centering
\caption{Attribute comparison of VSR works from 2011-2023: A Tabulated Overview}
	\label{Tab:Comparison2011}
	\resizebox{\linewidth}{!}{
	\begin{tabular}{|c|p{20mm}|p{15mm}|p{20mm}|p{15mm}|p{20mm}|p{15mm}|}
	\hline
	\centering \textbf{Existing Works}& 
	\centering \textbf{Holistic Feature Representation}& 
	\centering \textbf{Speaker Independence}& 
	\centering \textbf{Edge Device Deployment}& 
	\centering \textbf{Real-time Testing}&
    \centering \textbf{Benchmark Dataset Evaluation}&
    \centering \textbf{Accuracy $\geq$ 80\%}
\tabularnewline
\hline

\centering Damien et al. \cite{damien2011visual} &
\centering $\textcolor{green}{\checkmark}$ &
\centering $\textcolor{green}{\checkmark}$ &
\centering $\textcolor{red}{\times}$ &
\centering $\textcolor{green}{\checkmark}$ &
\centering $\textcolor{red}{\times}$ &
\centering $\textcolor{green}{\checkmark}$  \tabularnewline
\hline

\centering Shaik et al. \cite{shaikh2011visual} &
\centering $\textcolor{green}{\checkmark}$ &
\centering $\textcolor{green}{\checkmark}$ &
\centering $\textcolor{red}{\times}$ &
\centering $\textcolor{green}{\checkmark}$ &
\centering $\textcolor{red}{\times}$ &
\centering $\textcolor{green}{\checkmark}$  \tabularnewline
\hline

Sujatha et al. \cite{6179154} & 
\centering $\textcolor{green}{\checkmark}$ & 
\centering $\textcolor{green}{\checkmark}$ & 
\centering $\textcolor{red}{\times}$ & 
\centering $\textcolor{green}{\checkmark}$ &
\centering $\textcolor{red}{\times}$ & 
\centering $\textcolor{green}{\checkmark}$
\tabularnewline
\hline

Salah W et al. \cite{DBLP:journals/corr/abs-1301-4558}
& \centering $\textcolor{green}{\checkmark}$
& \centering $\textcolor{green}{\checkmark}$
& \centering $\textcolor{red}{\times}$
& \centering $\textcolor{green}{\checkmark}$
& \centering $\textcolor{red}{\times}$
& \centering $\textcolor{red}{\times}$ 
\tabularnewline
\hline

Helen L et al. \cite{7025274}
& \centering $\textcolor{green}{\checkmark}$
& \centering $\textcolor{green}{\checkmark}$
& \centering $\textcolor{red}{\times}$
& \centering $\textcolor{red}{\times}$
& \centering $\textcolor{red}{\times}$
& \centering $\textcolor{red}{\times}$ 
\tabularnewline
\hline

A.Amit et al. \cite{Amit2016LipRU}
& \centering $\textcolor{green}{\checkmark}$
& \centering $\textcolor{green}{\checkmark}$
& \centering $\textcolor{red}{\times}$
& \centering $\textcolor{red}{\times}$
& \centering $\textcolor{green}{\checkmark}$
& \centering $\textcolor{red}{\times}$ 
\tabularnewline
\hline

J. S. Chung et al. \cite{DBLP:journals/corr/ChungSVZ16}
& \centering $\textcolor{green}{\checkmark}$
& \centering $\textcolor{green}{\checkmark}$
& \centering $\textcolor{red}{\times}$
& \centering $\textcolor{red}{\times}$
& \centering $\textcolor{green}{\checkmark}$
& \centering $\textcolor{red}{\times}$ 
\tabularnewline
\hline

M. Assael et al. \cite{DBLP:journals/corr/AssaelSWF16}
& \centering $\textcolor{green}{\checkmark}$
& \centering $\textcolor{green}{\checkmark}$
& \centering $\textcolor{red}{\times}$
& \centering $\textcolor{red}{\times}$
& \centering $\textcolor{green}{\checkmark}$
& \centering $\textcolor{green}{\checkmark}$
\tabularnewline
\hline

Petridis et al. \cite{7472088}
& \centering $\textcolor{green}{\checkmark}$
& \centering $\textcolor{green}{\checkmark}$
& \centering $\textcolor{red}{\times}$
& \centering $\textcolor{red}{\times}$
& \centering $\textcolor{green}{\checkmark}$
& \centering $\textcolor{green}{\checkmark}$
\tabularnewline
\hline

Philip et al. \cite{8356433}
& \centering $\textcolor{green}{\checkmark}$
& \centering $\textcolor{green}{\checkmark}$
& \centering $\textcolor{red}{\times}$
& \centering $\textcolor{red}{\times}$
& \centering $\textcolor{green}{\checkmark}$
& \centering $\textcolor{green}{\checkmark}$
\tabularnewline
\hline

Eric Tatulli et al. \cite{7952701}
& \centering $\textcolor{green}{\checkmark}$
& \centering $\textcolor{green}{\checkmark}$
& \centering $\textcolor{red}{\times}$
& \centering $\textcolor{red}{\times}$
& \centering $\textcolor{red}{\times}$
& \centering $\textcolor{green}{\checkmark}$
\tabularnewline
\hline

P. Sindhura \cite{9001505} 
& \centering $\textcolor{red}{\times}$
& \centering $\textcolor{green}{\checkmark}$
& \centering $\textcolor{red}{\times}$
& \centering $\textcolor{red}{\times}$
& \centering $\textcolor{green}{\checkmark}$
& \centering $\textcolor{red}{\times}$
\tabularnewline
\hline

H. Gupta et al. \cite{8530509}
& \centering $\textcolor{red}{\times}$
& \centering $\textcolor{green}{\checkmark}$
& \centering $\textcolor{red}{\times}$
& \centering $\textcolor{red}{\times}$
& \centering $\textcolor{green}{\checkmark}$
& \centering $\textcolor{red}{\times}$
\tabularnewline
\hline

M. A. Abrar et al. \cite{9042439} & 
\centering $\textcolor{green}{\checkmark}$&
\centering $\textcolor{green}{\checkmark}$ &
\centering $\textcolor{green}{\checkmark}$&
\centering $\textcolor{green}{\checkmark}$
& \centering $\textcolor{green}{\checkmark}$
& \centering $\textcolor{red}{\times}$
\tabularnewline 
\hline

D. Parekh et al. \cite{9033664} &
\centering $\textcolor{green}{\checkmark}$ &
\centering $\textcolor{green}{\checkmark}$ 
& \centering $\textcolor{red}{\times}$ 
& \centering $\textcolor{red}{\times}$ 
& \centering $\textcolor{green}{\checkmark}$
& \centering $\textcolor{red}{\times}$
\tabularnewline  
\hline

Y.Lu et al. \cite{app9081599}
& \centering $\textcolor{red}{\times}$
& \centering $\textcolor{green}{\checkmark}$
& \centering $\textcolor{red}{\times}$
& \centering $\textcolor{red}{\times}$
& \centering $\textcolor{red}{\times}$
& \centering $\textcolor{green}{\checkmark}$
\tabularnewline
\hline

Guan et al. \cite{8929014}
& \centering $\textcolor{green}{\checkmark}$
& \centering $\textcolor{green}{\checkmark}$
& \centering $\textcolor{red}{\times}$
& \centering $\textcolor{green}{\checkmark}$
& \centering $\textcolor{red}{\times}$
& \centering $\textcolor{green}{\checkmark}$
\tabularnewline
\hline

N. K. Mudaliar et al. \cite{9137926}
& \centering $\textcolor{green}{\checkmark}$
& \centering $\textcolor{green}{\checkmark}$
& \centering $\textcolor{red}{\times}$
& \centering $\textcolor{green}{\checkmark}$
& \centering $\textcolor{green}{\checkmark}$
& \centering $\textcolor{green}{\checkmark}$
\tabularnewline
\hline

Yang et al. \cite{9190780}
& \centering $\textcolor{green}{\checkmark}$
& \centering $\textcolor{green}{\checkmark}$
& \centering $\textcolor{red}{\times}$
& \centering $\textcolor{red}{\times}$
& \centering $\textcolor{green}{\checkmark}$
& \centering $\textcolor{red}{\times}$
\tabularnewline
\hline

Qun Zhan et al. \cite{9506396}
& \centering $\textcolor{green}{\checkmark}$
& \centering $\textcolor{green}{\checkmark}$
& \centering $\textcolor{red}{\times}$
& \centering $\textcolor{red}{\times}$
& \centering $\textcolor{green}{\checkmark}$
& \centering $\textcolor{red}{\times}$
\tabularnewline
\hline

Xue et al. \cite{xue2023lipformer} 
& \centering $\textcolor{green}{\checkmark}$
& \centering $\textcolor{green}{\checkmark}$
& \centering $\textcolor{red}{\times}$
& \centering $\textcolor{green}{\checkmark}$
& \centering $\textcolor{green}{\checkmark}$
& \centering $\textcolor{red}{\times}$
\tabularnewline 
\hline

Ma et al. \cite{ma2022visual} 
& \centering $\textcolor{green}{\checkmark}$
& \centering $\textcolor{green}{\checkmark}$
& \centering $\textcolor{red}{\times}$
& \centering $\textcolor{red}{\times}$
& \centering $\textcolor{green}{\checkmark}$
& \centering $\textcolor{red}{\times}$
\tabularnewline 
\hline

Kim et al. \cite{kim2023prompt} 
& \centering $\textcolor{green}{\checkmark}$
& \centering $\textcolor{green}{\checkmark}$
& \centering $\textcolor{red}{\times}$
& \centering $\textcolor{green}{\checkmark}$
& \centering $\textcolor{green}{\checkmark}$
& \centering $\textcolor{red}{\times}$
\tabularnewline 
\hline

Nemani et al. \cite{nemani2022deep}
& \centering $\textcolor{green}{\checkmark}$
& \centering $\textcolor{green}{\checkmark}$
& \centering $\textcolor{green}{\checkmark}$
& \centering $\textcolor{green}{\checkmark}$
& \centering $\textcolor{green}{\checkmark}$
& \centering $\textcolor{green}{\checkmark}$
\tabularnewline 
\hline
\end{tabular}}\\
\end{table*}

\subsection{Speaker Independence}
Speaker independence is a critical aspect of VSR systems, as it enables them to recognize speech from various speakers without any prior training or adaptation. To achieve speaker independence, the VSR system must be trained on a large and diverse dataset including speech samples from many speakers. The feature extraction and normalization techniques used in the system must be robust to variations in speech patterns across different speakers. Additionally, the classification algorithm used in the VSR system must be able to generalize well to new speakers and not overfit to the training data from specific speakers. Finally, the VSR system should be evaluated using a testing dataset that includes speech samples from different speakers than the training dataset to ensure that the system can generalize to new speakers. By meeting these conditions, a speaker-independent VSR system can accurately recognize speech from unseen speakers. As demonstrated in Table \ref{Tab:Comparison2011}, the proposed solutions in the field of VSR for the period of 2011-2023 exhibit the concept of speaker independence to a great extent. Despite some limitations, this makes them highly relevant since they are based on datasets that support the concept of speaker independence. Notably, achieving speaker independence has been a significant challenge in the field of VSR, and the works that fulfill this requirement can be considered state-of-the-art in this domain.

\subsection{Effect of Content-Type on Speaker Independent VSR Model Performance}
The content of what is being said, such as digits, alphabets, or sentences, can significantly impact the performance of a speaker independent VSR system. For instance, recognizing digits and alphabets is generally considered easier than recognizing sentences due to the smaller vocabulary size and simpler context. This is because there are only ten digits and twenty-six alphabets to recognize, whereas recognizing sentences requires understanding the context of the sentence, which can be challenging. Additionally, the number of syllables in a word or sentence, the variation in pronunciation, and the complexity of the language can also impact the accuracy of VSR models. In speaker-independent VSR, dataset structure and variation are crucial factors that can significantly impact the performance of the model. A well-structured dataset can improve the model's accuracy and help it generalize better to new speakers and environments. Additionally, variations in the dataset, such as differences in lighting, background, and camera angles, can help the model learn to recognize visual speech patterns more effectively. Furthermore, VSR models may have a bias towards certain phonemes or words due to variations in pronunciation and the speaker's accent \cite{movellan1994visual, bertelson1997auditory}. For instance, certain words may be pronounced differently in different regions, leading to variations in phoneme recognition. Additionally, the speaker's speaking style may affect the model's performance, such as speaking rate, volume, and enunciation. Hence, it is essential to consider the variation in speaking styles and accents while training VSR models to achieve optimal performance. In conclusion, the content of what is being spoken can significantly affect the performance of a speaker independent VSR model. The complexity of the language, variations in pronunciation, and the speaker's accent and speaking style can all impact the model's accuracy. Therefore, developing VSR models that can handle these variations is crucial to achieving optimal performance.

\subsection{Edge Device Deployment and Real-time Testing in VSR systems}

\textbf{Edge device deployment} \cite{alajlan2022tinyml} and \textbf{real-time testing} are important aspects of VSR systems. Edge devices are small computing devices that can perform data processing and analysis close to the source of data collection \cite{cao2020overview, varghese2016challenges, shi2016edge}. In VSR, edge device deployment refers to the process of deploying the VSR system on small, portable, and low-power devices like mobile phones, tablets, or smart glasses. Real-time testing refers to the process of testing the VSR system while it is running in real-time, i.e., the system must process and classify the visual speech input within a very short time frame, typically a few hundred milliseconds, to be useful for practical applications. This requires the VSR system to be optimized for real-time performance, using efficient algorithms and architectures to quickly process the input. Edge device deployment and real-time testing are particularly important for VSR systems that are designed for applications in which the system must operate in real-time and in a distributed environment. For example, a VSR system deployed on smart glasses can provide real-time VSR for hearing-impaired individuals, enabling them to communicate in noisy environments or without attracting attention. Most existing works lack implemented solutions or do not rely on sensor infrastructure to accomplish VSR tasks. The edge device should be self-sufficient in predicting the output and should not require additional equipment. This would become a significant advantage of solutions that do not have a good user interface. Out of the solutions surveyed above, the presence of a user interface can be detected only in the solutions proposed by M. A. Abrar et al. \cite{9042439} and Nemani et al. \cite{nemani2022deep}. Though the solution portrays less testing accuracy, it forms the basis for many solutions that involve the presence of a UI.

\section{Conclusion}

This research provides a comprehensive overview of the methodologies employed in speaker-independent Visual Speech Recognition (VSR) and highlights the effectiveness of modern deep learning (DL) techniques compared to traditional approaches. The study evaluates the proposed solutions based on various parameters, ultimately concluding that a combination of convolutional neural networks (CNNs), recurrent neural networks (RNNs), and long short-term memory (LSTM) networks can effectively learn relevant features comparable to the traditional hand-crafted features reported in the existing literature. Moving forward, future research and development in speaker-independent VSR systems encompass several key areas. One such area is the exploration of transfer learning, which involves leveraging knowledge from pre-trained models and applying it to new models in different domains. This approach can help reduce the reliance on extensive training data, which is often limited in the speaker-independent VSR domain. Another area of interest is the investigation of unsupervised learning techniques, which can automatically discover data patterns without needing labeled training data. Leveraging unsupervised learning can be particularly beneficial in scenarios where large amounts of unlabeled data are available for speaker-independent VSR systems. The development of more sophisticated and robust deep learning models is another important focus for future research. Attention-based models, capable of handling varying lengths of input data and extracting relevant features more effectively, hold promise in advancing speaker-independent VSR. Furthermore, the exploration of new modalities, such as depth cameras and 3D imaging sensors, can contribute to building more robust and accurate VSR systems.

Integrating multiple modalities, such as audio and visual data, is an intriguing direction for enhancing the accuracy of speaker-independent VSR systems. The potential for improved performance and robustness increases by leveraging complementary information from different modalities. Moreover, the development of novel evaluation metrics that encompass multimodality and consider various aspects of VSR can provide more comprehensive assessments of system performance. In line with the evolving technological landscape, ensuring speaker-independent VSR systems' security is of utmost importance. Research efforts should be directed toward investigating adversarial attacks and other security issues to develop countermeasures and safeguard the integrity and reliability of VSR systems. In summary, this research highlights the efficacy of modern DL techniques in speaker-independent VSR and identifies various areas for future exploration. The proposed avenues, including transfer learning, unsupervised learning, advanced deep learning models, new modalities, multimodal integration, novel evaluation metrics, and security considerations, pave the way for advancing the field and enhancing the capabilities of speaker-independent VSR systems.

\section{Acknowledgements}
The authors would like to express their sincere gratitude to IIIT Naya Raipur for providing the necessary resources and support to carry out this research on speaker-independent VSR. The authors would also like to thank all the researchers worldwide who have dedicated their time and effort toward improving the field of speaker-independent VSR. Their research and contributions have helped to lay the foundation for the work presented in this study.

\section{Abbreviations}

We have included a table in which the abbreviations used throughout this document are defined for the convenience of the reader. This table aims to clearly understand the abbreviations used and prevent any confusion or misunderstanding while reading the document. By referring to this table, readers can quickly and easily find the meaning of any abbreviation that may be unfamiliar to them, thereby enhancing their overall comprehension of the text. We believe that providing this table will help readers navigate the content easily and improve their reading experience.

\begin{table}[htbp]

	\centering
	\caption{Abbreviations}
 \label{Tab: abb}
	\resizebox{\linewidth}{!}{\begin{tabular}{|c|c|}
	\hline
 
 \centering \textbf{Abbreviation} & 
 \centering \textbf{Meaning} \tabularnewline 
 \hline

 \centering VSR &
 \centering Visual Speech Recognition \tabularnewline 
 \hline

 \centering DCT &
 \centering Discrete Cosine Transform \tabularnewline
 \hline

 \centering PCA &
 \centering Principal Component Analysis \tabularnewline
 \hline

 \centering DWT &
 \centering Discrete Wavelet Transform \tabularnewline
 \hline

\centering HMM &
 \centering Hidden Markov Model \tabularnewline 
 \hline

 \centering SVM &
 \centering Support Vector Machine \tabularnewline 
 \hline

 \centering CNN &
 \centering Convolutional Neural Network \tabularnewline 
 \hline

 \centering RNN &
 \centering Recurrent Neural Network  \tabularnewline 
 \hline

 \centering LSTM &
 \centering Long short-term memory  \tabularnewline 
 \hline

 \centering STCNN &
 \centering Spatio-Temporal Convolutional Neural Network \tabularnewline
 \hline

 \centering MHIs &
 \centering Motion History Images \tabularnewline
 \hline

  \centering EM &
 \centering Expectation-Maximization \tabularnewline
 \hline
 
  \centering GA &
 \centering Genetic Algorithm \tabularnewline
 \hline

   \centering VGG &
 \centering Visual Geometry Group \tabularnewline
 \hline

    \centering KNNs &
 \centering K- Nearest Neigbours \tabularnewline
 \hline
 
	\end{tabular}}
\end{table}

\bibliographystyle{IEEEtran}
\bibliography{LIPAR}
\end{document}